\documentclass[10pt,twocolumn,letterpaper]{article}

\usepackage[ready]{iccv}   
\usepackage{multirow}
\usepackage{multicol}
\usepackage{makecell}
\usepackage[accsupp]{axessibility}  

\definecolor{iccvblue}{rgb}{0.21,0.49,0.74}
\usepackage[pagebackref,breaklinks,colorlinks,allcolors=iccvblue]{hyperref}

\def\paperID{} 
\def\confName{ICCV-CV4E}
\def\confYear{2025}

\title{ForestFormer3D: A Unified Framework for End-to-End Segmentation of Forest LiDAR 3D Point Clouds}

\author{
Binbin Xiang$^{1}$, Maciej Wielgosz$^{1}$, Stefano Puliti$^{1}$, Kamil Král$^{2}$\\
Martin Krůček$^{2}$, Azim Missarov$^{2}$, Rasmus Astrup$^{1}$ \\
$^{1}$Norwegian Institute of Bioeconomy Research (NIBIO)\\
$^{2}$Silva Tarouca Research Institute for Landscape and Ornamental Gardening\\
{\tt\small \{binbin.xiang, maciej.wielgosz, stefano.puliti, rasmus.astrup\}@nibio.no}\\
{\tt\small \{kral, krucek, missarov\}@vukoz.cz}
}

\begin{document}
\maketitle
\begin{abstract}
The segmentation of forest LiDAR 3D point clouds, including both individual tree and semantic segmentation, is fundamental for advancing forest management and ecological research. However, current approaches often struggle with the complexity and variability of natural forest environments. We present ForestFormer3D, a new unified and end-to-end framework designed for precise individual tree and semantic segmentation. ForestFormer3D incorporates ISA-guided query point selection, a score-based block merging strategy during inference, and a one-to-many association mechanism for effective training. By combining these new components, our model achieves state-of-the-art performance for individual tree segmentation on the newly introduced FOR-instanceV2 dataset, which spans diverse forest types and regions. Additionally, ForestFormer3D generalizes well to unseen test sets (Wytham woods and LAUTx), showcasing its robustness across different forest conditions and sensor modalities. The FOR-instanceV2 dataset and the ForestFormer3D code are publicly available at \url{https://bxiang233.github.io/FF3D/}.
\end{abstract}    
\section{Introduction}
\label{sec:intro}

\begin{figure*}[!htbp]
  \captionsetup{skip=0pt} 
  \centering
  \vspace*{-12pt} 
  \includegraphics[width=\textwidth]{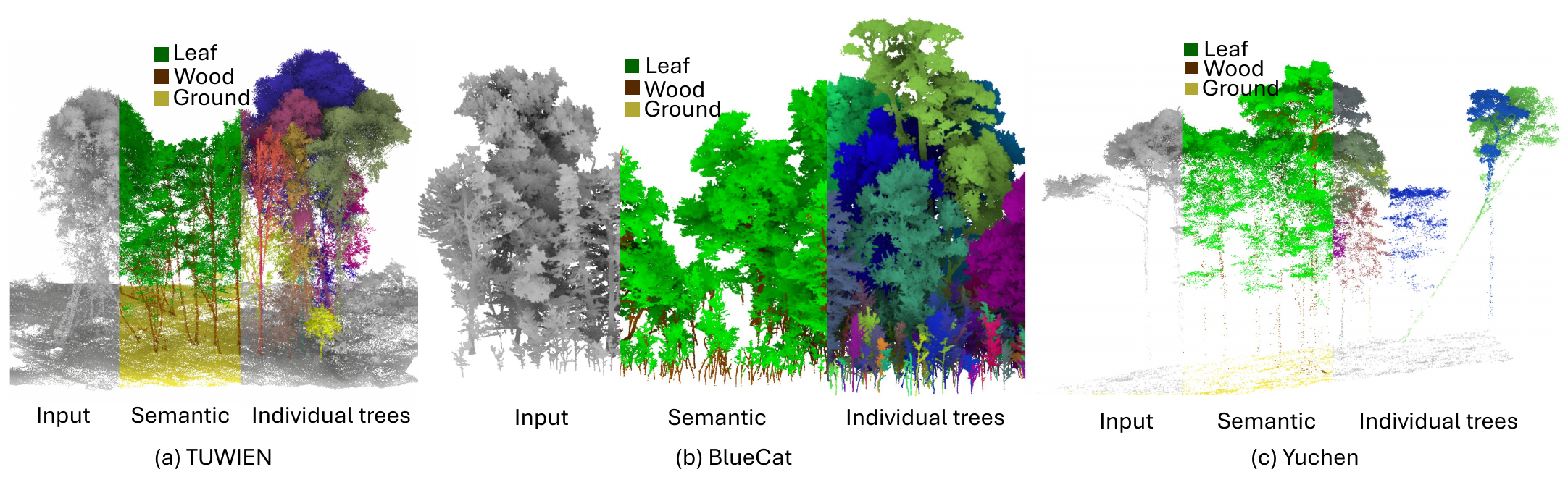} 
  \caption{Examples of semantic and individual tree segmentation in different forest regions from our dataset, visualized with consistent point size. The colors for the individual trees are assigned randomly. The figure highlights the density and distribution variability between regions (a)–(c). (a) TUWIEN: Deciduous-dominated alluvial forest in Austria, showing dense, interwoven tree structures. (b) BlueCat: This area includes both large and small trees, highlighting the challenge of segmenting trees of significantly varied sizes within the same region. (c) Yuchen: A tropical forest region with tall trees and sparse points in the lower stem regions due to occlusion, adding difficulty to segmentation tasks, particularly for stem part.}
  \label{fig:challenges}
  \vspace*{-12pt} 
\end{figure*}

Forests, covering approximately 31\% of the Earth’s land surface, are essential for biodiversity conservation, carbon storage, provision of timber and a wide range of ecosystem services central to the quality of life. Forests are complex 3D structures and to fully understand and map the forest ecosystems requires a detailed 3D representation of the structure. Advancements in high-density LiDAR technology now enable large-scale capture of 3D point clouds with unprecedented detail, paving the way for improved and streamlined forest ecosystem data capture. However, to capitalize on the captured data, further advancement in semantic and instance segmentation of 3D forest point clouds is still required. Significant progress has been made in 3D segmentation for indoor and urban environments, but moving to forest scenes introduces a substantial increase in complexity. The organic, intertwined structure of trees, highly variable sizes and shapes, and intricate occlusion patterns demand specialized segmentation methods. In simpler forest structures, like boreal forests, LiDAR-derived point clouds have been shown to enable tasks such as forest semantic segmentation, tree instance segmentation, and tree biophysical parameter estimation~\cite{Xiang2024Automated}. However, in more complex broadleaved, temperate, or tropical forests, increased structural complexity and species richness demand specialized and novel segmentation methods beyond current approaches~\cite{WIELGOSZ2024114367}, which can address the following key unresolved challenges:

\begin{itemize}
    \item \textbf{Structural complexity.} Dense canopies, characterized by closely spaced trees with plastic and overlapping tree crowns, complicate the separation of individual trees (see \cref{fig:challenges}(a)).
    \item \textbf{Multi-scale representation.} Forests contain both large canopy trees and small understory trees, requiring models capable of segmenting across multiple scales (see \cref{fig:challenges}(b)).
    \item \textbf{Variability in point distribution.} Differences in acquisition sensors, forest types, and environmental conditions lead to variations in point density and distribution, necessitating robust generalization across diverse forest scenes and sensor configurations (see differences among \cref{fig:challenges}(a)-(c)).
\end{itemize}

Focusing on improving the state-of-the-art for more complex forests, this paper provides two primary contributions:

\begin{itemize}
    \item \textbf{FOR-instanceV2 dataset}: We introduce an expanded dataset based on FOR-instance~\cite{Puliti2023Forinstance}, with a primary focus on increasing forest structural complexity by incorporating broadleaved temperate and tropical forests. This dataset encompasses a wider range of forest types and regions, establishing a robust benchmark for forest point cloud segmentation tasks. This dataset includes dedicated training, validation, and test sets, enabling comprehensive evaluation of segmentation accuracy and robustness.

    \item \textbf{ForestFormer3D}: We propose ForestFormer3D, a unified, end-to-end framework for individual tree segmentation and semantic segmentation of forest point clouds. ForestFormer3D achieves state-of-the-art performance for individual tree segmentation on the FOR-instanceV2 test set and demonstrates strong generalization capabilities on two previously unseen test regions (Wytham woods and LAUTx), highlighting its adaptability and effectiveness across diverse forest types.
\end{itemize}


\section{Related work}
\label{sec:relatedwork}

\textbf{3D point cloud segmentation.}  
3D point cloud semantic and instance segmentation are fundamental tasks in 3D computer vision. Semantic segmentation assigns each point in a point cloud to a predefined category, typically supervised by cross-entropy loss and developed through advancements in network architectures and computational strategies~\cite{Qi2017PointNet,Qi2017aPointNet,Thomas2019KPConv,Choy20194D}. Instance segmentation extends this by assigning a unique instance label to each point, allowing for the distinction of individual objects within each category.

Much research in 3D point cloud segmentation has focused on indoor scenes, with benchmarks such as ScanNet~\cite{Dai2017ScanNet} and S3DIS~\cite{Armeni20163D} commonly used for evaluation. Outdoor segmentation has primarily targeted applications in autonomous driving, utilizing datasets like SemanticKITTI~\cite{Behley2019SemanticKITTI} and nuScenes~\cite{Caesar2019nuScenes}, which focus on traffic-related classes such as pedestrians and vehicles. However, these existing methods and datasets are optimized for structured urban environments, leaving more complex, unstructured natural 3D scenes, such as forests, comparatively underexplored.

Existing deep learning approaches for instance segmentation in 3D point clouds can be divided into three main types: top-down, bottom-up and transformer-based methods. Top-down approaches typically detect objects first, often using bounding boxes or proposals, and then apply binary mask prediction within these detected regions~\cite{Yi2019GSPN,Yang2019Learning,Kolodiazhnyi2023TopDown,Zhang2020Instance}. These methods rely on the initial detection stage to localize objects accurately. Bottom-up approaches focus on extracting discriminative point-level features, which are then clustered into instances~\cite{Wang2018SGPN,Liu2019MASC,Pham2019JSIS3D,Elich20193D-BEVIS,Han2020Occuseg,Engelmann20203D-MPA,Jiang2020PointGroup,Chen2021Hierarchical,Vu2022SoftGroup,Zhong2022MaskGroup}. These methods are capable of capturing fine-grained details but are prone to issues such as over-segmentation or under-segmentation. Moreover, their reliance on clustering hyperparameters introduces additional challenges in tuning these methods for complex, variable-density point clouds. 

Transformer-based approaches have recently emerged as a promising solution, offering fully end-to-end frameworks that leverage attention mechanisms to model complex spatial relationships while avoiding error accumulation through intermediate stages~\cite{Schult2023Mask3d,Sun2023Superpoint,Kolodiazhnyi2024OneFormer3D}. This unified design allows for joint training of both semantic and instance segmentation tasks, potentially enhancing performance through mutual learning effects~\cite{Kolodiazhnyi2024OneFormer3D}. In these architectures, a 3D backbone such as sparse 3D U-Net~\cite{Choy20194D} extracts global features from the point cloud, which are processed by a transformer decoder with a fixed set of queries to concurrently predict semantic and instance masks~\cite{Schult2023Mask3d,Sun2023Superpoint,Kolodiazhnyi2024OneFormer3D}. Recent research has further optimized this framework, focusing on query initialization and refinement techniques to improve alignment between instance queries and object locations, thereby reducing the need for extensive iterative refinement~\cite{Lu2023Query,Sun2023Superpoint}. Additionally, integrating geometric and semantic information has proven effective in enhancing mask quality and boundary precision by addressing limitations in spatial cues and fine-grained detail capture during query refinement~\cite{Yao2024SGIFormer,Shin2024Spherical}. To handle multi-scale representations and maintain spatial consistency in large scenes, techniques capturing both local details and global context have been introduced, further advancing segmentation performance in diverse and complex environments~\cite{Liu20223DQueryIS,Loiseau2024Learnable}.

\noindent\textbf{Forest 3D scene segmentation.}
Forest point cloud segmentation has long been a significant challenge~\cite{hyppa2001segmentation,Sorin2022Estimating} aimed primarily at localizing and segmenting individual tree crowns~\cite{Antonio20123d,Dalponte2016,Ayrey2017,Liu2021Individual,Wilkes2023TLS2trees}, and tree semantic segmentation, distinguishing for example leaf from wood~\cite{Wang2019LeWoS,Vicari2019,Wan2021Anovel}. These methods often rely on density-based techniques, hand-crafted features, or canopy height models (CHMs), which are effective in certain controlled settings but lack adaptability across varied forest types, requiring extensive manual adjustments and showing poor generalizability.

Accurate instance and semantic segmentation of forest point clouds could fundamentally streamline numerous forestry tasks, transforming individual tree detection, species classification, and biophysical parameter prediction into by-products of a comprehensive segmentation model. The growing availability of high-density LiDAR data, along with open labelled datasets~\cite{calders2022laser,tockner_andreas_2022_6560112,Puliti2023Forinstance,NEURIPS2023_9708c7d3,WIELGOSZ2024114367}, provide the foundation necessary for deep learning methods to achieve automated and adaptable forest point cloud segmentation~\cite{Xi2018Aug2Filtering,Chen2021Jan21Individual,Krisanski2021Apr7Sensor,Sun2022Jun14Individual,Chang2022Oct6Two,Wang2023Feb13Tree,Zhang2023Jun5Towards,Kim2023Jun5Automated,Jiang2023Jun25LWSNet,Wielgosz2023Jul27P2T,Straker2023Instance,Zhou2025TreeStructor}. However, progress in forest segmentation lags behind developments in urban and indoor contexts. The majority of forestry research has focused on semantic segmentation tasks, such as separating leaf and wood points~\cite{Xi2018Aug2Filtering,Krisanski2021Apr7Sensor,Kim2023Jun5Automated,Jiang2023Jun25LWSNet}. Few studies focus on individual tree segmentation, with most relying on multi-step, hand-crafted pipelines that limit their applicability across diverse forest types and sensor setups~\cite{Windrim2020May6Detection, Wielgosz2023Jul27P2T}.

To bridge this gap, recent frameworks aim to integrate multiple segmentation tasks into a single model~\cite{Xiang2024Automated,WIELGOSZ2024114367,Henrich2024TreeLearn}. For example, the ForAINet framework combines semantic and instance segmentation using a bottom-up approach~\cite{Xiang2024Automated}. While effective in most simple forest environments, this model tends to over-segment trees in densely packed areas and fails to accurately identify small trees in multi-layered forests. 

In this paper, we present ForestFormer3D, the first end-to-end framework for individual tree segmentation and semantic segmentation tailored specifically for forest point clouds. Unlike previous bottom-up methods that rely on non-differentiable clustering steps for individual tree segmentation~\cite{Xiang2024Automated,WIELGOSZ2024114367,Henrich2024TreeLearn}, our transformer-based decoder enables fully differentiable, end-to-end training. Moreover, ForestFormer3D eliminates the need for intermediate matching by directly supervising predicted masks with their corresponding ground truth instances. This approach offers both a streamlined architecture and improved individual tree segmentation accuracy. The key innovations (see \cref{fig:framework}) in our framework include:

\begin{itemize}
    \item We introduce both an instance- and a semantic-aware guided (ISA-guided) query point selection strategy to generate high-quality query points, helping in balancing precision and recall in individual tree segmentation and ensuring more complete instance coverage.
    \item Leveraging the known positions of query points, we apply a one-to-many association method for efficient mask supervision, replacing the Hungarian-based one-to-one matching strategy commonly adopted in earlier methods.
    \item To process large-scale forest point clouds, we implement a score-based global ranking block-merging strategy, allowing seamless integration across extensive scenes.
\end{itemize}

\section{Method}

\begin{figure*}
  \centering
  \vspace*{-16pt}
  \includegraphics[width=\textwidth]{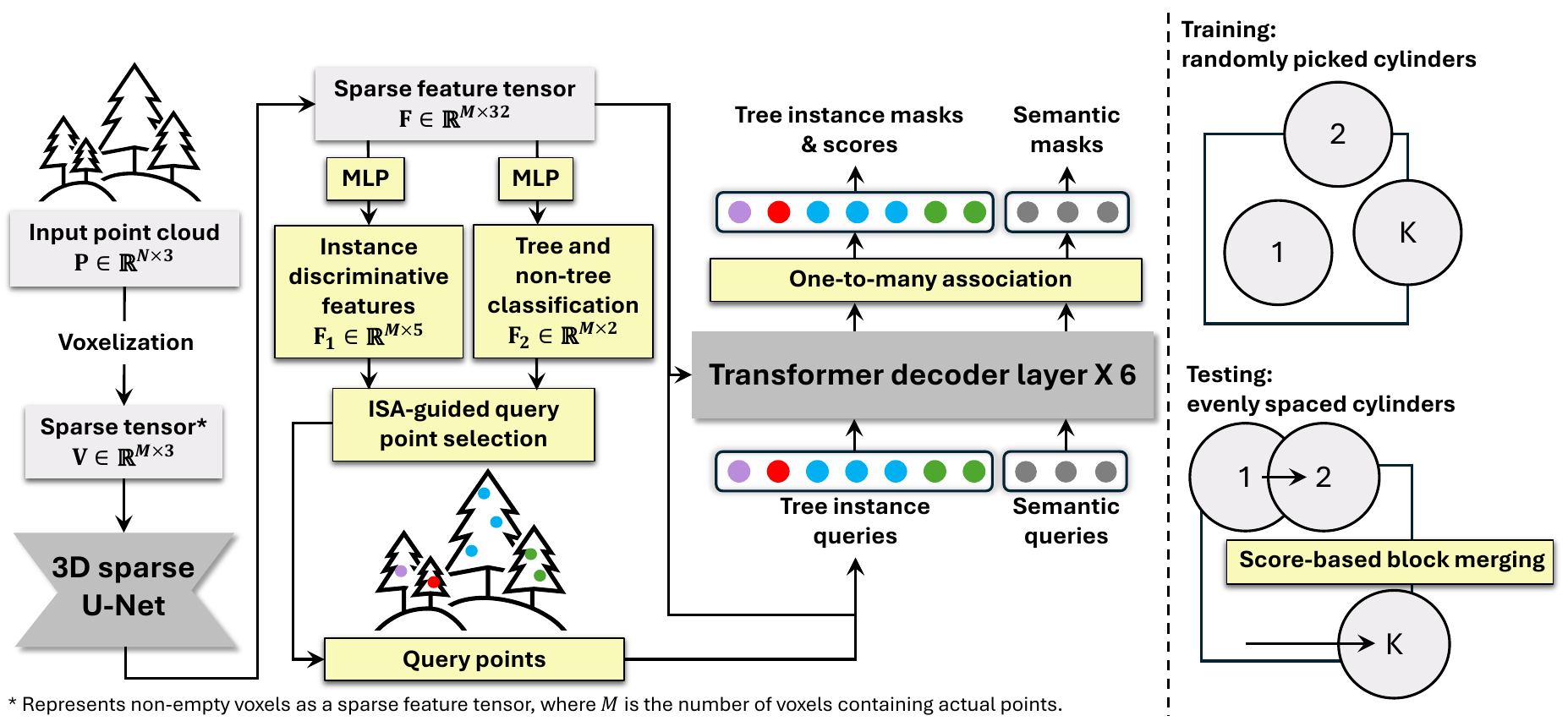}
  \caption{Illustration of our ForestFormer3D framework, with key innovative components highlighted in yellow.}
  \label{fig:framework}
  \vspace*{-14pt}
\end{figure*}

Our method aims at the combined task of semantic and individual tree segmentation in forest point clouds. The general framework of ForestFormer3D is shown in \cref{fig:framework}. The semantic segmentation involves three primary classes in forest environments: ground, wood and leaf. Unlike indoor datasets, which commonly include RGB and mesh information~\cite{Armeni2017Joint,Dai2017ScanNet}, forest point clouds are usually captured by LiDAR. Although LiDAR data often provides additional information such as intensity, number of returns, and return numbers, these attributes are not consistently available across all datasets. Therefore, to ensure our method is broadly applicable, we use only the 3D coordinates.

At first, we voxelize the input point cloud and create a sparse tensor. Then, we process this tensor through a sparse 3D U-Net~\cite{Choy20194D} to extract voxel-wise features. These features are used for both instance and semantic segmentation. We use a novel ISA-guided sampling strategy to select high-quality query points for individual tree segmentation. The selected instance query points, along with the semantic query points, are fed into the query decoder. For each query point, the decoder outputs a corresponding mask: instance query points produce individual tree masks, while semantic query points generate semantic region masks (see \cref{sec:network_architecture}).

Since the ISA-guided sampling strategy retains the spatial information of the query points, we can directly associate each predicted mask with its corresponding tree instance, eliminating the need for one-to-one matching algorithms like the Hungarian algorithm~\cite{Kuhn1955The,Schult2023Mask3d,Lu2023Query,Kolodiazhnyi2024OneFormer3D}. This direct association allows us to efficiently compute the mask loss, enabling the network to perform both semantic segmentation and individual tree segmentation in an end-to-end manner (see \cref{sec:training}).

Given the large-area coverage of point clouds in forest scenes, we adopt a cropping strategy to handle large inputs. Specifically, we crop the input point cloud into cylindrical regions with a certain radius, processing one local block at a time, as shown on the right side of \cref{fig:framework}. We use cylindrical blocks to avoid cutting trees vertically and to improve neighbor search efficiency~\cite{Xiang2024Automated}. During training, we use random cropping. During testing, we predict inside each cylinder block individually, then apply a sliding window approach with a fixed stride to cover the entire point cloud.  Finally, we propose a new score-based block merging method to handle overlapping regions, merging all blocks into a complete point cloud prediction (see \cref{sec:inference}).

\subsection{Network architecture}
\label{sec:network_architecture}

\textbf{3D sparse U-Net backbone.} Given an input point cloud \( \mathbf{P} \in \mathbb{R}^{N \times 3} \), with \( N \) representing the number of 3D points, we voxelize \( \mathbf{P} \) to obtain a sparse tensor \( \mathbf{V} \in \mathbb{R}^{M \times 3} \), where \( M \) is the count of occupied voxels. Using the SpConvUNet~\cite{Contributors2022Spconv}, we extract features from \( \mathbf{V} \), outputting a feature tensor \( \mathbf{F} \in \mathbb{R}^{M \times 32} \).

\noindent\textbf{ISA-guided query point selection.} Previous research has explored various methods for query point selection in instance segmentation tasks, as the quality of the query points critically impacts both model performance and convergence speed. Some approaches use learnable tensors as queries, which are typically initialized randomly or with zero values~\cite{Kolodiazhnyi2024OneFormer3D}. While these parametric queries offer flexibility, they converge slowly~\cite{Meng2021Conditional} and can be challenging to control or interpret visually, as their initialization and refinement are purely data-driven. Other methods, such as farthest point sampling (FPS), select non-parametric queries directly from the raw input~\cite{Schult2023Mask3d}. Non-parametric approaches are efficient but may produce suboptimal query points that overlook smaller instances or sample points in uninformative background regions.

To address these limitations, recent strategies have proposed more controllable query point selection methods. For example, semantic-aware approaches prioritize selecting query points from foreground regions to avoid generating false positive masks from background classes~\cite{He2023FastInst}. Other methods incorporate learning-based query selection to dynamically refine query points. These approaches typically generate initial query points and then refine their relationships or employ clustering techniques to improve the query distribution~\cite{Lu2023Query,Liu20223DQueryIS}.

We develop a novel ISA-guided query point selection strategy to improve the quality of the selected query points, minimizing the likelihood of sampling background points and ensuring more uniform and high-coverage sampling of tree instances. Using the output tensor \( \mathbf{F} \in \mathbb{R}^{M \times 32} \) from the 3D sparse U-Net, we process it through two parallel Multi-Layer Perceptron (MLP) heads. The first head learns a 5-dimensional (5D) embedding for each voxel, such that voxels belonging to the same tree are closer in the embedding space, while voxels from different trees are pushed farther apart. This separation is achieved using an instance discriminative loss to encourage compactness within trees and separation between trees~\cite{DeBrabandere2017Semantic,Wang2019Associatively,Xiang2023Towards}. The second MLP head performs binary classification to separate tree voxels from non-tree voxels. During instance query point sampling, we exclude non-tree voxels and apply FPS in the 5D embedding space of tree voxels to select a fixed number of instance query points.


We conducted a small study to evaluate the effectiveness of our sampling method on the FOR-instanceV2 validation set (see \cref{sec:experimentalsettings}). Using a sliding window approach, we sampled 300 query points within each cylindrical region of radius \(R=16\,\text{m}\) and a stride of \( R/4 \). For each cylinder, we calculated the coverage rate as the proportion of tree instances represented by selected query points and the proportion of selected query points corresponding to tree voxels. Averaging these values across all cylinders, our ISA-guided query point selection achieved a coverage rate of \( 92.7\% \) and a tree voxel ratio of \( 98.5\% \). In comparison, simple FPS-based sampling~\cite{Schult2023Mask3d} yielded a coverage rate of \( 79.9\% \) and a tree voxel ratio of \( 81.3\% \). These results demonstrate that our approach maximizes coverage of all tree instances while minimizing the sampling of ground voxels.

\noindent\textbf{Query decoder.} The query decoder uses \( K_{\text{ins}} \) instance query points selected by our ISA-guided method and \( K_{\text{sem}} \) learnable semantic queries (\( K_{\text{sem}} = 3 \) in our case). These semantic queries are randomly initialized and updated during training to capture semantic information. The sparse feature tensor \( \mathbf{F} \in \mathbb{R}^{M \times 32} \) serves as the key and value. Following previous methods~\cite{Sun2023Superpoint, Kolodiazhnyi2024OneFormer3D}, the decoder applies six transformer layers with self-attention on the queries and cross-attention with keys and values, outputting \( K_{\text{ins}} \) tree instance masks, \( K_{\text{sem}} \) semantic masks, and confidence score from 0 to 1 for each predicted tree instance mask.

\subsection{Training}
\label{sec:training}


\noindent\textbf{One-to-many association.} Our method explicitly selects instance query points during the query point selection step, meaning their spatial positions are inherently determined by the selection process. Since these queries are pre-aligned with tree instances at selection time, each predicted tree instance mask can be directly associated with its corresponding ground truth (GT) instance, eliminating the need for an additional matching step. In contrast, most existing methods~\cite{Kolodiazhnyi2024OneFormer3D,Kuhn1955The,Schult2023Mask3d} require optimization-based matching to establish correspondences between predicted masks and GT instances. Inspired by one-to-many assignment strategies~\cite{Zong2023DETRs}, we calculate score and mask losses for all predicted masks.

In previous one-to-one matching setups~\cite{Kolodiazhnyi2024OneFormer3D,Kuhn1955The,Schult2023Mask3d}, the goal is to ensure that each GT instance mask matches with only one optimal predicted mask, enforcing distinct relationships among queries or masks and suppressing redundant masks. However, we observe that one-to-one matching, as in OneFormer3D, often produces numerous false positives during inference, leading to lower final precision for individual tree segmentation and still requiring an additional matrix-NMS (non-maximum suppression) step to remove duplicate masks~\cite{Kolodiazhnyi2024OneFormer3D}.

In contrast, our approach leverages a more controlled instance query selection strategy, which aims to cover all tree instances effectively, maximizing recall. With accurate score predictions, our method quickly filters out duplicate masks, thereby improving precision as well. This combination results in higher-quality individual tree predictions overall. For semantic segmentation, we simplify the association process: given the three semantic masks, we directly associate them with the GT masks in order.

\noindent\textbf{Loss.} The instance-related loss consists of binary cross-entropy loss \( \mathcal{L}_{\text{bce}} \), Dice loss \( \mathcal{L}_{\text{dice}} \), and score loss \( \mathcal{L}_{\text{score}} \)~\cite{Kolodiazhnyi2024OneFormer3D}. The score loss is computed using mean squared error, with the target score defined as the intersection-over-union (IoU) between each predicted mask and the GT mask that has the highest IoU with it. The total instance loss is:

\vspace{-4pt} 
\begin{equation}
  \vspace*{-3pt} 
  \mathcal{L}_{\text{instance}} = \mathcal{L}_{\text{bce}} + \mathcal{L}_{\text{dice}} + 0.5 \cdot \mathcal{L}_{\text{score}}.
  \vspace*{-3pt} 
  \label{eq:loss_instance}
\end{equation}
\vspace{-4pt} 

The semantic loss \( \mathcal{L}_{\text{sem}} \) is the cross-entropy loss~\cite{Kolodiazhnyi2024OneFormer3D}. Since our decoder consists of six layers, each layer’s output is compared with the GT to calculate the instance and semantic losses, and these are summed across all layers:

\vspace{-4pt}
\begin{equation}
\vspace*{-3pt} 
\mathcal{L}_{\text{total}} = \sum_{\text{layer}=1}^{6} (\mathcal{L}_{\text{instance}} + 0.2 \cdot \mathcal{L}_{\text{sem}}),
\vspace*{-1pt} 
\label{eq:loss_total}
\end{equation}
\vspace{-4pt}

where the weights for loss terms are as in~\cite{Sun2023Superpoint, Kolodiazhnyi2024OneFormer3D}. Additionally, we include the losses from the two MLP heads introduced in ISA-guided query point selection (\cref{sec:network_architecture}): the binary cross-entropy loss \( \mathcal{L}_{\text{binary}} \), and the discriminative loss \( \mathcal{L}_{\text{disc}} \)~\cite{DeBrabandere2017Semantic,Wang2019Associatively,Xiang2023Towards}. The final complete loss is thus:

\vspace{-4pt}
\begin{equation}
\vspace*{-3pt} 
\mathcal{L} = \mathcal{L}_{\text{total}} + \mathcal{L}_{\text{binary}} + \mathcal{L}_{\text{disc}}.
\vspace*{-3pt} 
\label{eq:loss_final}
\end{equation}
\vspace{-4pt}

\subsection{Inference}
\label{sec:inference}

\textbf{Score-based block merging.} In forest point clouds, cropping is unavoidable due to the large spatial extent, which necessitates a merging strategy to handle overlapping regions (see \cref{fig:framework}). For semantic segmentation, merging is relatively simple, as predictions can be resolved by voting across overlapping regions. However, block merging strategies for instance segmentation remain limited and face several challenges.

Most existing instance segmentation tasks focus on indoor datasets (where blocks often represent individual rooms) or outdoor datasets for autonomous driving (where blocks correspond to individual scans). In these settings, block merging is unnecessary and therefore rarely studied. In contrast, forest datasets cover vast areas, requiring a sliding window approach with fixed step sizes and overlapping regions to manage computational constraints.

Existing block merging methods~\cite{Wang2018SGPN,Denis2023Improved} rely primarily on rule-based strategies, setting overlap thresholds to determine whether instances in neighboring blocks should merge. In our approach, we use a score-based method: leveraging the scores obtained from the query decoder for each predicted mask, we rank all masks across the scene and apply NMS to remove lower-scoring, overlapping masks.

To further address issues caused by cylinder cropping (which may split individual trees), inspired by the approach of Henrich \etal~\cite{Henrich2024TreeLearn}, we discard masks near the crop boundary. Specifically, for a cylinder radius of 16\,\text{m}, we remove any predicted instance whose mask includes even a single point within 0.5\,\text{m} of the cylinder boundary.


\section{Experiments}

\subsection{Experimental settings}
\label{sec:experimentalsettings}

\noindent\textbf{Dataset.} We extend the FOR-instance dataset~\cite{Puliti2023Forinstance} by incorporating additional challenging datasets, resulting in a new, comprehensive dataset named FOR-instanceV2. The newly added datasets include NIBIO2, NIBIO\_MLS, BlueCat, and Yuchen regions. These additions increase the dataset’s diversity and segmentation complexity by including Mobile Laser Scanning (MLS) data (NIBIO\_MLS) collected via backpack LiDAR, data from tropical regions (Yuchen), and regions with highly varied tree sizes (NIBIO2 and BlueCat), where both large and small trees coexist within the same area. In addition, BlueCat (6.3k instances) and NIBIO2 (3k instances) are previously unavailable datasets, publicly released for the first time. They extend FOR-instance (1.1k instances) eightfold number of trees. This expanded dataset provides a more challenging benchmark for point cloud segmentation in forests, and we plan to release FOR-instanceV2 as an open dataset to support future research and foster advancements in the field.

\begin{figure}[!htbp]
  \captionsetup{skip=0pt} 
  \centering
  \includegraphics[width=\columnwidth]{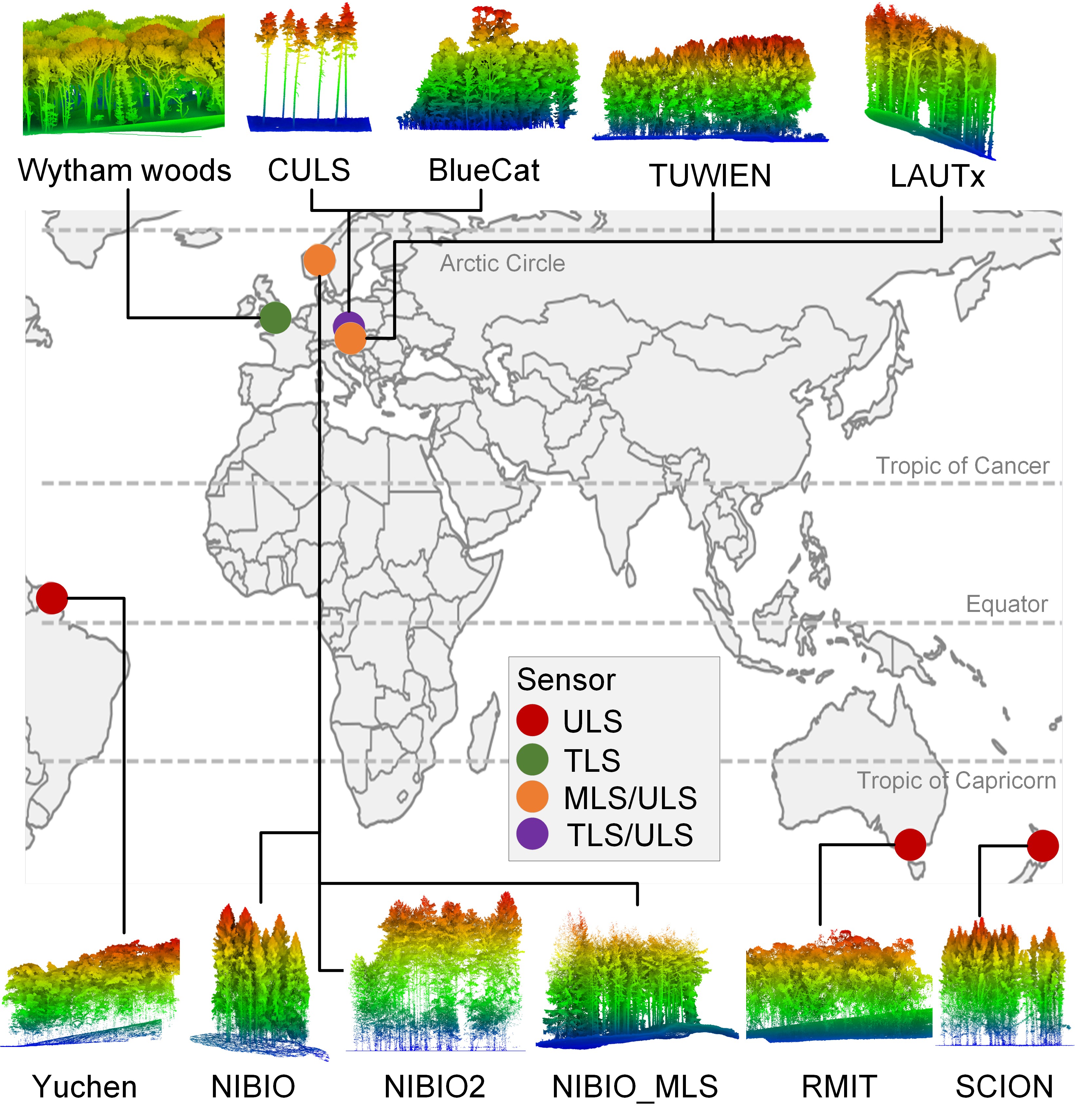}
  \caption{Geographical overview of the FOR-instanceV2 dataset and the two additional test datasets, LAUTx and Wytham woods. Different colored markers indicate the sensors used for data collection.}
  \label{fig:datasetmap}
  \vspace*{-4pt} 
\end{figure}

To assess transferability, we also include two additional datasets, LAUTx and Wytham woods, which are used exclusively for testing and do not contain any training or validation data. \cref{fig:datasetmap} illustrates a geographical overview of FOR-instanceV2 and these two test datasets. We provide detailed statistics and annotation protocols in the supplementary materials.

\noindent\textbf{Metrics.} We adopt precision (Prec), recall (Rec), and F1-score (F1) for individual tree segmentation. Predicted tree instances are matched to GT trees based on the IoU of their respective point sets, with predictions below an IoU threshold of 0.5 considered false positives. To further evaluate the delineation accuracy of tree instances, we include a coverage metric (Cov) that quantifies the alignment between predicted and GT boundaries~\cite{Wang2019Associatively,Xiang2024Automated}. For each GT tree, the predicted instance with the highest IoU is identified, and coverage is calculated as the average IoU across all GT trees. The mean intersection-over-union (mIoU) is used as the semantic segmentation metric.

\subsection{Comparison to baselines}
\label{sec:comparisontobaselines}

All baseline methods and our proposed ForestFormer3D were trained on the FOR-instanceV2 training set and evaluated on the corresponding test split, as shown in \cref{tab:BaselineCompare}. ForAINet~\cite{Xiang2024Automated} and TreeLearn~\cite{Henrich2024TreeLearn} are two recent and effective deep learning-based methods for forest point cloud segmentation. ForAINetV2\_R8 maintains the original configuration of ForAINet with an input cylinder radius of 8\,\text{m}, while ForAINetV2\_R16 uses a 16\,\text{m} radius to match ForestFormer3D and isolate the effect of input size. TreeLearn~\cite{Henrich2024TreeLearn}, designed for ground-based LiDAR forest point clouds, predicts offsets from individual points to their respective tree bases, defined as the trunk location three meters above ground. We adapted the training parameters of TreeLearn for our experiments, reducing the epochs from 1400 to 1200, the learning rate from 0.002 to 0.001, and the total number of samples from 25,000 to 2,500, while increasing the batch size from 2 to 8 and adjusting the chunk size from 35 to 30. While effective for regions with dense stem points, this method struggles with ULS data (\eg, NIBIO2, RMIT and Yuchen), where stem points are sparse, showing its poor generalizability across diverse sensor modalities. In particular, TreeLearn fails to predict individual trees in the Yuchen region due to the extreme sparsity of stem points. As a result, we compare TreeLearn’s performance only on regions excluding Yuchen in the FOR-instanceV2 test split (as shown in \cref{tab:BaselineCompare}). OneFormer3D~\cite{Kolodiazhnyi2024OneFormer3D}, a state-of-the-art method in 3D point cloud panoptic segmentation, is adapted for forest point cloud segmentation by employing the same block merging strategy as ForAINet, using overlap thresholds to determine whether instances in neighboring blocks should merge.

The results in \cref{tab:BaselineCompare} demonstrate that ForestFormer3D outperforms all baselines, achieving the highest F1, surpassing the second-best method by 9.7 percent points (pp), and the highest Cov, exceeding the next best by 1.2\,pp. This reflects ForestFormer3D's superior ability to produce complete and accurate individual tree segmentations. While all methods provide unique instance IDs for each point, the high Prec of ForestFormer3D suggests a reduced likelihood of over-segmentation compared to baselines.

\begin{table}[h]
\centering
\vspace*{-6pt} 
\caption{Comparison of individual tree segmentation and semantic segmentation results with baselines in FOR-instanceV2 test split. The best results are in bold, and the second best ones are \underline{underlined}.}
\vspace*{-6pt} 
\label{tab:BaselineCompare}
\scriptsize
\begin{tabular}{@{}l@{\hskip 4pt}c@{\hskip 4pt}c@{\hskip 4pt}c@{\hskip 4pt}c@{\hskip 4pt}c@{\hskip 4pt}c@{\hskip 4pt}c@{\hskip 4pt}c@{}}
\toprule
\multirow{2}{*}{Method} & \multicolumn{4}{c}{Individual Tree Seg. (\%)} & \multicolumn{4}{c}{Semantic Seg. (\%)} \\ 
\cmidrule(l{-1pt}r{4pt}){2-5} \cmidrule(l{1pt}r{0pt}){6-9}
 & Prec & Rec & F1 & Cov & Ground & Wood & Leaf & mIoU \\ \midrule
ForAINetV2\_R8 & 84.3 & 63.4 & 72.4 & 73.3 & \underline{98.3} & \underline{68.1} & \underline{94.3} & \underline{86.9} \\
ForAINetV2\_R16 & \underline{88.1} &  59.2& 70.8 & 72.7 & \textbf{98.8} & \textbf{69.2} & \textbf{94.5} & \textbf{87.5} \\
TreeLearn~\cite{Henrich2024TreeLearn} & 82.0 & 36.6 & 50.6 & 52.2 & -- &--  & -- & -- \\
OneFormer3D~\cite{Kolodiazhnyi2024OneFormer3D} & 72.0 & \underline{74.3} & \underline{73.1} & \underline{80.0} & 97.8 & 63.1 & 93.4 & 84.8 \\
ForestFormer3D (ours) &\textbf{92.4} & \textbf{75.0} & \textbf{82.8} & \textbf{81.2} & 96.9 & 67.7 & 94.1 & 86.2 \\ \bottomrule
\end{tabular}
\vspace*{-6pt} 
\end{table}

To evaluate the transferability of our model, we test on Wytham woods~\cite{calders2022laser} and LAUTx~\cite{tockner_andreas_2022_6560112} datasets, which represent mixed-species and natural broadleaved temperate forests with diverse structural complexity. As shown in \cref{tab:compareOtherdataset}, ForestFormer3D achieves best performance of individual tree segmentation compared to previous methods evaluated on the Wytham woods dataset, achieving the highest F1 and Cov. Although TreeLearn~\cite{Henrich2024TreeLearn}, a method designed for ground-based LiDAR data, achieves the best results on the LAUTx dataset, ForestFormer3D still ranks second in F1 and Cov, and maintains competitive semantic segmentation performance. These results demonstrate the model’s good generalization ability to previously unseen forest types. Notably, the Wytham woods region includes tree species not present in the FOR-instanceV2 training set, while the LAUTx dataset, collected using MLS, further validates the adaptability of ForestFormer3D across different sensor modalities and forest structures.

\begin{figure*}[]
  \captionsetup{skip=0pt} 
  \vspace*{-7pt} 
  \centering
  \includegraphics[width=0.85\textwidth]{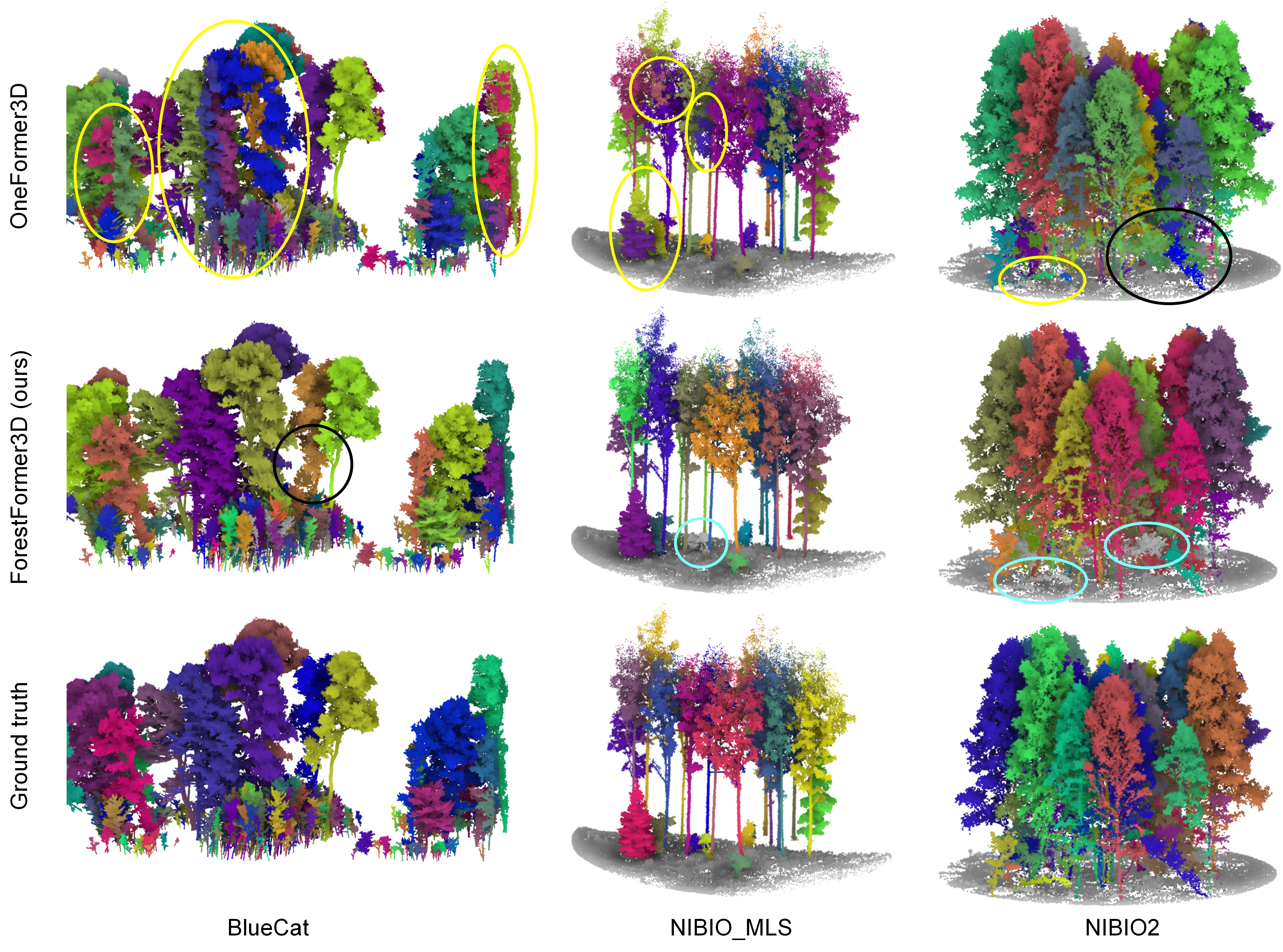}
  \caption{Visual comparison of individual tree segmentation results between ForestFormer3D and OneFormer3D, with randomly assigned colors representing different tree instances. Segmentation errors are highlighted using three distinct markers: yellow circles for over-segmentation, bright blue circles for undetected trees, and black circles for wrongly assigning points from one tree to a neighboring tree.} \label{fig:comparewithotherbaselines_main}
  \vspace*{-10pt} 
\end{figure*}

\cref{fig:comparewithotherbaselines_main} presents the qualitative results of ForestFormer3D and OneFormer3D for individual tree segmentation across different regions, demonstrating ForestFormer3D’s ability to effectively mitigate over-segmentation and produce more precise mask predictions.

Across datasets, our method achieves substantial F1 improvements for individual tree segmentation (+9.7\,pp on FOR-instanceV2 and +8.2\,pp on Wytham woods), with marginal mIoU reductions for semantic segmentation (-1.3\,pp and -1.0\,pp, respectively). While mainly optimized for individual tree segmentation, our framework also performs semantic segmentation within the same unified structure, leading to a slight reduction in mIoU, likely due to shared representation learning prioritizing instance-level distinctions. However, the overall semantic segmentation performance remains stable.

\begin{table}[h]
\centering
\vspace*{-10pt} 
\caption{Comparison with previous methods on the individual tree segmentation and semantic segmentation for Wytham woods and LAUTx test data. The best results are in bold, and the second best ones are \underline{underlined}.}
\vspace*{-5pt} 
\label{tab:compareOtherdataset}
\scriptsize
\begin{tabular}{@{}p{0.5cm} p{2cm} p{0.2cm} p{0.15cm} p{0.15cm} p{0.15cm} p{0.4cm} p{0.2cm} p{0.15cm}@{}}
\toprule
\multirow{2}{*}{\makecell[tl]{Region}} & \multirow{2}{*}{\makecell[tl]{Method}} & \multicolumn{4}{c}{Individual Tree Seg. (\%)} & \multicolumn{3}{c}{Semantic Seg. (\%)} \\ 
\cmidrule(l){3-6} \cmidrule(l){7-9} 
                       &                     & Prec & Rec & F1 & Cov & Ground & Tree & mIoU \\ \midrule
\multirow{6}{*}{\makecell[tl]{Wytham \\ woods\\ ~\cite{calders2022laser}}}  
& ForAINetV2\_R16 &  \textbf{82.2}   &  53.8   & 65.1    &   54.4   &84.2  &  \underline{99.8}    & 92.0 \\
& SegmentAnyTree~\cite{WIELGOSZ2024114367}       & 73.0  & 53.0 & 61.4 & -- & -- & -- & -- \\  & TreeLearn~\cite{Henrich2024TreeLearn} &  52.5   &  \textbf{91.9}   & \underline{66.8}    &   49.2   &--  &  --    & --\\
                       & OneFormer3D~\cite{Kolodiazhnyi2024OneFormer3D} &  41.4    &  62.4   & 49.7    &   \underline{61.6}   &  \textbf{88.8}    &   \textbf{99.9}   &  \textbf{94.4}\\ 
                       & ForestFormer3D (ours) &  \underline{81.8}   &  \underline{69.2}   & \textbf{75.0}    &   \textbf{66.9}   &\underline{86.9}  &  \textbf{99.9}    & \underline{93.4} \\ \midrule
\multirow{8}{*}{\makecell[tl]{LAUTx \\ ~\cite{tockner_andreas_2022_6560112}}} 
& Point2tree~\cite{Wielgosz2023Jul27P2T}       & 85.0  & 55.0 & 66.8 & -- & -- & -- & -- \\ 
& TLS2trees~\cite{Wilkes2023TLS2trees}       & 70.0  & 40.5 & 51.3 & -- & -- & -- & -- \\ 
& ForAINetV2\_R16       &93.7  & 77.7 & 85.0 & 74.7 & \textbf{95.9} & \underline{96.7} & \textbf{96.3} \\ 
& SegmentAnyTree~\cite{WIELGOSZ2024114367}       & 91.0  & 75.0 & 82.2 & -- & -- & -- & -- \\ & TreeLearn~\cite{Henrich2024TreeLearn}      & \textbf{96.8} &   \textbf{87.9} &   \textbf{92.1} &\textbf{85.2} & 92.6 & 95.9 & 94.3 \\
                       & OneFormer3D~\cite{Kolodiazhnyi2024OneFormer3D} &   58.1   &  80.2   &  67.4   &   74.6   &  \underline{94.6}    &   \textbf{96.9}   &  \underline{95.8}\\ 
                       & ForestFormer3D (ours) &   \underline{95.7}   &   \underline{85.9}  & \underline{90.5}    &  \underline{79.1}    & 94.1     &   \underline{96.7}   &  95.4\\ 
\bottomrule
\end{tabular}
\vspace*{-15pt} 
\end{table}

\subsection{Ablation studies}
\label{sec:ablationstudies}

We conduct a series of ablation studies to analyze the impact of the three proposed key components of ForestFormer3D: ISA-guided query point selection, the score-based block merging strategy during inference, and one-to-many association during training (see \cref{tab:ablationstudiesfornetworkstructure}). Here, we fix the score threshold on mask removal to 0.6 for all methods. Additionally, we examine the effect of other factors on the performance of ForestFormer3D in the supplementary materials.

\noindent\textbf{ISA-guided query point selection.} As shown in \cref{tab:ablationstudiesfornetworkstructure}, the ISA-guided query point selection improves the F1 on the FOR-instanceV2 test set by 3.4\,pp and increases Prec by 7.8\,pp compared to the baseline OneFormer3D, which uses randomly generated learnable query points. This strategy ensures high-quality query points that are both instance-discriminative (ensuring high individual tree coverage) and semantically aware (avoiding picking non-tree points), thereby reducing the generation of false positive masks. On the validation set, the ISA-guided query point selection improves the F1 score by 0.9\,pp and Prec by 4.4\,pp compared to the baseline.

\noindent\textbf{Score-based block merging strategy.} For the block merging strategy, two key designs contribute to the F1 improvement (see \cref{tab:ablationstudiesfornetworkstructure}). First, replacing the overlap-based merging strategy of the baseline with a score-based approach (BM1) to remove overlapping instances results in an F1 increase of 3.1\,pp on the test split. When ISA-guided query point selection and BM1 are combined, the F1 improves by 5.4\,pp compared to the baseline. The second enhancement involves discarding predicted masks near crop boundaries to address inaccuracies caused by cropping (BM2). When combined with ISA-guided query point selection, BM2 improves the F1 by an additional 2.2\,pp over BM1. A similar pattern can also be observed in the validation split.

\noindent\textbf{One-to-many association.} Changing the one-to-one matching strategy in OneFormer3D to our one-to-many association yields a further 0.8\,pp increase on both the FOR-instanceV2 test and validation set (see \cref{tab:ablationstudiesfornetworkstructure}). 

Overall, incorporating all three key components results in an 8.4\,pp increase in F1, a 0.4\,pp improvement in Cov, and a 1.5\,pp increase in mIoU compared to the baseline in test split. A similar pattern can also be observed in the validation split, except that the semantic segmentation mIoU dropped by 0.5\,pp compared to the baseline.

\begin{table}[ht]
\centering
\vspace*{-5pt} 
\caption{Ablation study for three proposed key components on the FOR-instanceV2 dataset. ISA represents ISA-guided query point selection, BM1 indicates score-based block merging without discarding masks near the crop boundary, BM2 uses score-based block merging with discarding masks near the crop boundary, and O2MA denotes one-to-many association. The best results are in bold, and the second best ones are \underline{underlined}.}
\vspace*{-5pt} 
\label{tab:ablationstudiesfornetworkstructure}
\scriptsize
\begin{tabular}{@{}c@{\hskip 4pt}c@{\hskip 4pt}c@{\hskip 4pt}c@{\hskip 4pt}c@{\hskip 4pt}c@{\hskip 4pt}c@{\hskip 4pt}c@{\hskip 4pt}c@{}}
\toprule
\multirow{2}{*}{ISA} & \multirow{2}{*}{BM1} & \multirow{2}{*}{BM2} & \multirow{2}{*}{O2MA} & \multicolumn{4}{c}{Individual Tree Seg. (\%)} & Semantic Seg. (\%) \\ 
\cmidrule(l{-1pt}r{3pt}){5-8}  \cmidrule(l{-1pt}r{0pt}){9-9} 
                     &                      &                      &                     & Prec & Rec & F1  & Cov  & mIoU \\ \midrule
\multicolumn{9}{l}{\textbf{Test split}}  \\ 
\multicolumn{4}{l}{OneFormer3D, baseline}  & 72.0 & \textbf{74.3} & 73.1 & \underline{80.0} & 84.8 \\ 
\midrule
\checkmark          &                      &                      &                     & 80.2     &   73.1  &  76.5   &  \underline{80.0}    & 84.1     \\
    &     \checkmark             &                      &                     &  79.8    &  73.0   &  76.2   &   78.6   &  84.8   \\
     \checkmark    &     \checkmark             &                      &                     &  87.3    &   71.2  &  78.5   &  78.6    &  84.3    \\ 
   \checkmark      &                &       \checkmark                 &                     &  \underline{89.4}    &   \underline{73.5}  &  \underline{80.7}   &  79.3    &  \underline{85.2}    \\
   \checkmark      &               &   \checkmark           &   \checkmark                  &  \textbf{93.9}    &  71.9   &  \textbf{81.5}   &    \textbf{80.4}  &   \textbf{86.3}   \\\midrule
\multicolumn{9}{l}{\textbf{Validation split}}  \\ 
\multicolumn{4}{l}{OneFormer3D, baseline}  & 70.2 & \textbf{68.1} & 69.1 & \underline{80.2} & \textbf{84.1} \\ \midrule 
\checkmark          &                      &                      &                     & 74.6     &   66.0  &  70.0   &  79.4   & 82.4     \\
& \checkmark &                      &                     &  78.6    &  65.4   &  71.4   &   77.4   &  \textbf{84.1}   \\
 \checkmark & \checkmark &                      &                     &  84.6    &   64.6  &  73.2  &  78.3    &  82.6    \\ 
 \checkmark &       &   \checkmark           &                     &  \underline{86.9}    &   \underline{67.3}  &  \underline{75.9}   &  78.7    &  83.3   \\
 \checkmark &       &   \checkmark           &   \checkmark                  &  \textbf{93.2}    &  65.1   &  \textbf{76.7}   &    \textbf{80.5}  &   \underline{83.6}   \\\bottomrule
\end{tabular}
\vspace*{-10pt} 
\end{table}

\section{Conclusion}

We introduced ForestFormer3D, an end-to-end framework designed for individual tree segmentation and semantic segmentation for forest 3D LiDAR point clouds. It incorporates three innovative components: ISA-guided query point selection, a score-based block merging strategy for inference, and one-to-many association during training. ForestFormer3D achieves excellent individual tree segmentation performance on our newly introduced dataset (FOR-instanceV2) and generalizes well to two additional unseen datasets (Wytham woods and LAUTx). These results highlight its potential as a unified model, capable of effectively handling point clouds collected from diverse forest types, geographic regions, and sensor modalities without being constrained by specific conditions. 

We hope this work inspires research into segmentation methods tailored for forest environments and encourages the community to experiment and advance algorithms using the FOR-instanceV2 dataset. ForestFormer3D aims to serve as a strong baseline, facilitating further developments and practical applications in forestry analysis, environmental monitoring, and sustainable resource management.

\section*{Acknowledgements}

This work is part of the Center for Research-based Innovation SmartForest: Bringing Industry 4.0 to the Norwegian forest sector (NFR SFI project no.\,309671, smartforest.no). MK, KK, and AM were funded by the INTER-COST project LUC23023.


\clearpage
\setcounter{page}{1}
\maketitlesupplementary

\renewcommand{\thesection}{\arabic{section}}
\setcounter{section}{0}

\section{Overview} \label{sec:Overviewsupplementary}
In this supplementary material, we provide: 
\begin{itemize}
    \item implementation details (\cref{sec:supimplementationdetails}),
    \item additional dataset details and point cloud annotation (\cref{sec:supdatasetdetails}),
    \item a quantitative comparison including both individual tree segmentation and semantic segmentation across nine regions in the FOR-instanceV2 test split (\cref{sec:OtherComparisontobaselines}),
    \item additional ablation studies (\cref{sec:Otherablationstudies}),
    \item qualitative results (\cref{sec:Qualitativeresults}),
    and \item relation to forest domain specific metrics (\cref{sec:MetricRelation}).
\end{itemize}

\section{Implementation details} \label{sec:supimplementationdetails}

\begin{figure*}[htbp]
    \centering
    \includegraphics[width=\textwidth]{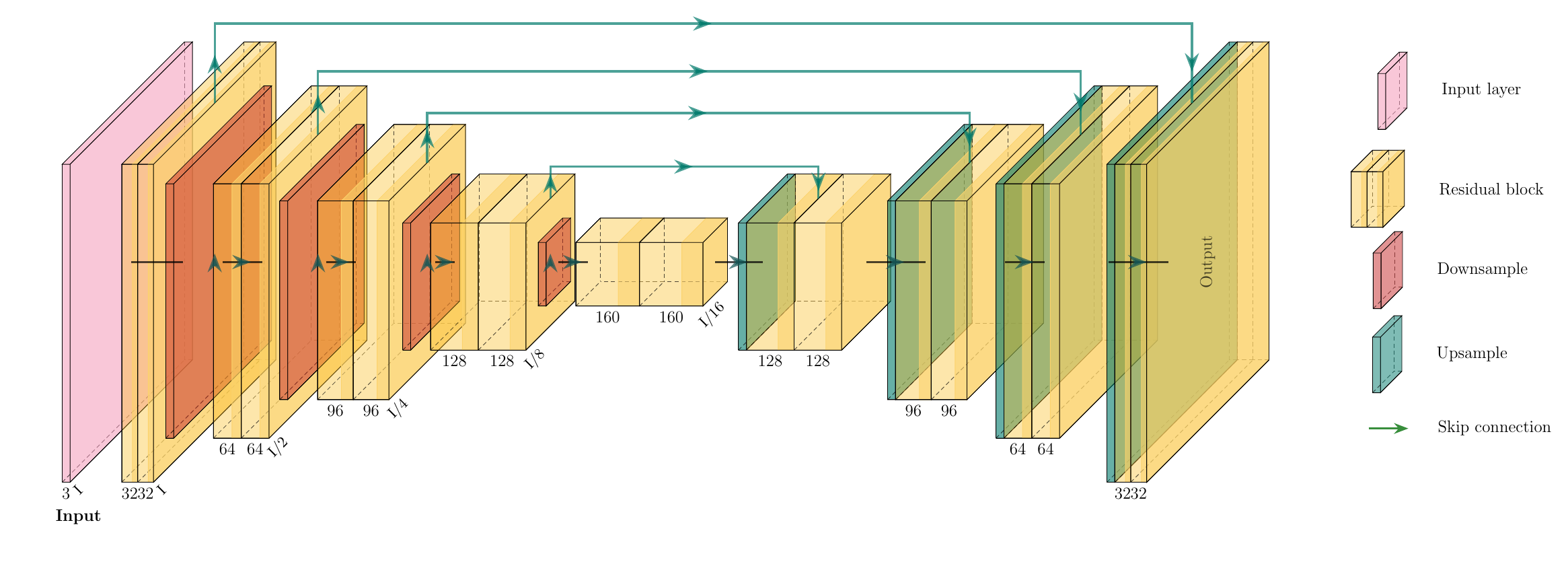}
    \caption{The detailed structure of the 3D sparse U-Net architecture used for feature extraction from the voxelized 3D point cloud.}
    \label{fig:spconvunet}
\end{figure*}

\noindent \textbf{3D sparse U-Net architecture.} The 3D sparse U-Net used for sparse tensor feature extraction is shown in~\cref{fig:spconvunet}. 

\noindent \textbf{Loss calculation.} All instance masks, predicted by transformer decoder, are supervised by losses including binary cross-entropy (BCE), Dice, and score loss.

\begin{equation}
\mathcal{L}_{\text{bce}} = - \frac{1}{N} \sum_{i=1}^{N} \left( y_i \log \sigma(\hat{y}_i) + (1 - y_i) \log (1 - \sigma(\hat{y}_i)) \right),
\label{eq:loss_bce}
\end{equation}

\begin{equation}
\mathcal{L}_{\text{dice}} = 1 - \frac{2 \sum_{i=1}^{N} \sigma(\hat{y}_i) y_i + 1}{\sum_{i=1}^{N} \sigma(\hat{y}_i) + \sum_{i=1}^{N} y_i + 1},
\label{eq:loss_dice}
\end{equation}

where $N$ is the number of voxels, \( \hat{y}_i \) is the predicted mask logits at voxel \( i \), \( y_i \in \{0,1\} \) is the GT mask value, and \( \sigma(\cdot) \) represents the sigmoid function. The additive constant \( +1 \) in ~\cref{eq:loss_dice} prevents division by zero.

The score loss supervises the predicted confidence scores of all predicted instance masks using mean squared error:

\begin{equation}
\mathcal{L}_{\text{score}} = \frac{1}{M} \sum_{j=1}^{M} (\hat{s}_j - s_j)^2,
\label{eq:loss_score}
\end{equation}

where $M$ is the number of predicted masks, \( \hat{s}_j \) is the predicted score for predicted mask \( j \), and \( s_j \) is set as the IoU between the predicted mask and its best matched GT mask, or zero if no GT overlaps. IoU is computed based on the number of intersecting voxels divided by the union of the predicted and GT mask voxels.

To facilitate query point selection, we first construct a 5D embedding space where voxels from the same instance are close while those from different instances are apart. This embedding is learned using discriminative loss:

\begin{equation}
\mathcal{L}_{\text{disc}} = \mathcal{L}_{\text{var}} + \mathcal{L}_{\text{dist}} + 0.001 \cdot \mathcal{L}_{\text{reg}}.
\label{eq:loss_disc}
\end{equation}

\( \mathcal{L}_{\text{var}} \) encourages intra-instance compactness by pulling voxel embeddings within the same instance toward their mean. \( \mathcal{L}_{\text{dist}} \) enforces inter-instance separation, and \( \mathcal{L}_{\text{reg}} \) constrains instance centroids to stay near the origin. Let \(C\) denote the total number of tree instances, and \( \mathcal{I}_c \) the set of voxels belonging to instance \( c \). Each voxel \( i \) has an embedding \( {f}_i \), and the mean embedding of instance \( c \) is \( \mu_c \). We define \( N_c\) as the number of voxels in instance \( c \). The margins \(\delta_v\) and \(\delta_d\) control the desired intra-instance compactness and inter-instance separation, respectively. Three loss terms are calculated by:

\begin{equation}
\mathcal{L}_{\text{var}} = \frac{1}{C} \sum_{c=1}^{C} \frac{1}{N_c} 
\sum_{i \in \mathcal{I}_c} 
\left[ \max \bigl(\|f_i - \mu_c\|_1 - \delta_v, 0\bigr) \right]^2,
\label{eq:loss_var_new}
\end{equation}

\begin{equation}
\mathcal{L}_{\text{dist}} = \frac{1}{C(C-1)} \sum_{c_1 \neq c_2} 
\left[ \max \bigl( 2\delta_d - \|\mu_{c_1} - \mu_{c_2}\|_1, 0 \bigr) \right]^2,
\label{eq:loss_dist_new}
\end{equation}

\begin{equation}
\mathcal{L}_{\text{reg}} = \frac{1}{C}\sum_{c=1}^{C} \|\mu_c\|_1.
\label{eq:loss_reg_new}
\end{equation}

We set \( \delta_v = 0.5 \) and \( \delta_d = 1.5 \) following prior work~\cite{DeBrabandere2017Semantic,Wang2019Associatively,Xiang2023Towards}.

In addition, to ensure that query points are sampled from tree voxels, we train a tree and non-tree classification head to distinguish tree and non-tree voxels using the BCE loss:



\begin{equation}
\mathcal{L}_{\text{binary}} = - \frac{1}{N} \sum_{i=1}^{N} ( l_i \log \hat{l}_i + (1 - l_i) \log (1 - \hat{l}_i) ),
\label{eq:loss_binary}
\end{equation}

\noindent

where \(l_i \) is the GT label for voxel \(i\), and \(\hat{l}_i \) is the predicted class probability.

\noindent\textbf{Other implementation details.} Point clouds are voxelized to a resolution of 0.2\,\text{m}. Data augmentation includes random horizontal flipping, random rotation around the Z\text{-axis}, and random scaling by a factor between 0.8 and 1.2. We train our model on a single NVIDIA A100 GPU with 80 GB memory, using a batch size of 2 for 3000 epochs. 

To stabilize training, we first pretrain two MLP heads introduced in ISA-guided query point selection for 1000 epochs before jointly training the full model. Specifically, the 5D embedding feature head is trained with the discriminative loss (see~\cref{eq:loss_disc}) to enforce compact intra-instance representations and inter-instance separation, while the binary classification head is trained with a binary cross-entropy loss to distinguish tree voxels from non-tree voxels (see~\cref{eq:loss_binary}). During this warm-up stage, the decoder remains frozen, ensuring that the MLP heads produce stable embeddings and classifications before integrating with the full network. After this stage, we jointly train all components, including the 3D sparse U-Net and the transformer decoder, optimizing all loss terms together. 

For inference, we sample 300 query points within each cylindrical region of radius 16\,\text{m} with a stride of 4\,\text{m} using a sliding window approach. The best score threshold on mask removal is 0.4 according to our ablation studies (\cref{fig:ablationscorethreshold}(d)).

\section{Dataset details and point cloud annotation} \label{sec:supdatasetdetails}

\cref{tab:ForInstanceDataInfos} summarizes the key characteristics of the FOR-instanceV2 dataset and the additional test datasets.

For the BlueCat data annotation, all trees were manually labeled using 3D Forest software~\cite{Trochta20173DForest} by two independent annotators working on separate, non-overlapping datasets. After labeling, the trees were matched to field measurements based on spatial proximity. This pairing was validated by comparing tree heights derived from the point cloud with those estimated from DBH-based allometry. In addition, every tree with a DBH of 1\,cm or greater was field-verified to eliminate false positives, and any obvious labeling errors were manually refined.

For the NIBIO2 data, annotations were performed using CloudCompare software (\href{https://www.cloudcompare.org}{https://www.cloudcompare.org}, V\,2.12.4) by two annotators, and the results were later reviewed to ensure consistency and accuracy.

\begin{table*}[h]
\centering
\caption{Characteristics of the FOR-instanceV2 dataset and additional test datasets in different geographic regions. The FOR-instanceV2 dataset contains unique individual tree IDs for each point, as well as semantic labels for ground, wood, and leaf. The additional test data, on the other hand, includes only tree and non-tree labels, along with individual tree IDs for each point.}
\label{tab:ForInstanceDataInfos}
\scriptsize
\begin{tabular}{
    p{0.1cm}    
    p{1.5cm}    
    p{6.0cm}    
    p{2.7cm}    
    >{\centering\arraybackslash}p{0.2cm}  
    >{\centering\arraybackslash}p{0.2cm}  
    >{\centering\arraybackslash}p{0.1cm}  
    p{0.6cm}    
    p{1.8cm}    
}
\toprule
 &
  \multirow{2}{*}{Region name} &
  \multirow{2}{*}{Forest type (Tree species)} &
  \multirow{2}{*}{Scanning mode (Sensor)} &
  \multicolumn{3}{c}{Number of plots} &
  \multirow{2}{*}{\begin{tabular}[c]{@{}l@{}}Number\\ of trees\end{tabular}} &
  \multirow{2}{*}{Country} \\ 
\cmidrule(lr){5-7}
     &    &                                              &                     & Train & Val & Test &     &                \\ 
\midrule
\multirow{12}{*}{\rotatebox{90}{\scriptsize FOR-instanceV2}} 
& CULS \cite{Puliti2023Forinstance} &
  Coniferous dominated temperate forest (\textit{Pinus sylvestris}) &
  ULS (Riegl VUX-1) &
  \phantom{0}1 & \phantom{0}1 & \phantom{0}1 & \phantom{00}37 & Czech Republic \\
& NIBIO \cite{Puliti2023Forinstance} &
  Coniferous dominated boreal forest (\textit{Picea abies}, \textit{Pinus sylvestris}, \textit{Betula sp.} (few)) &
  ULS (Riegl MiniVUX-1) &
  \phantom{0}8 & \phantom{0}6 & \phantom{0}6 & \phantom{0}575 & Norway \\
& RMIT \cite{Puliti2023Forinstance} &
  Native dry sclerophyll eucalypt forest (\textit{Eucalyptus sp.}) &
  ULS (Riegl MiniVUX-1) &
  \phantom{0}1 & \phantom{0}0 & \phantom{0}1 & \phantom{0}223 & Australia \\
& SCION \cite{Puliti2023Forinstance} &
  Non-native pure coniferous temperate forest (\textit{Pinus radiata}) &
  ULS (Riegl MiniVUX-1) &
  \phantom{0}2 & \phantom{0}1 & \phantom{0}2 & \phantom{0}135 & New Zealand \\
& TUWIEN \cite{Puliti2023Forinstance} &
  Deciduous dominated temperate forest (Deciduous species) &
  ULS (Riegl VUX-1) &
  \phantom{0}1 & \phantom{0}0 & \phantom{0}1 & \phantom{0}150 & Austria \\ 
& NIBIO2 &
  Coniferous dominated boreal forest (\textit{Pinus sylvestris}, \textit{Picea abies}, \textit{Betula sp.}) &
  ULS (Riegl VUX-1) &
  29 & \phantom{0}6 & 15 & 3062 & Norway \\
& NIBIO\_MLS \cite{WIELGOSZ2024114367} &
  Coniferous dominated boreal forest (\textit{Picea abies}, \textit{Pinus sylvestris}, \textit{Betula sp.}) &
  MLS (GeoSLAM ZEB-HORIZON) &
  \phantom{0}4 & \phantom{0}1 & \phantom{0}1 & \phantom{0}258 & Norway \\
& BlueCat &
  Deciduous temperate forest (Mixed species) &
  TLS (Leica P20) &
  \phantom{0}1 & \phantom{0}1 & \phantom{0}1 & 6304 & Czech Republic \\ 
& Yuchen \citep{NEURIPS2023_9708c7d3} &
  Tropical forest &
  ULS (Riegl MiniVUX-1) &
  \phantom{0}1 & \phantom{0}1 & \phantom{0}1 & \phantom{0}281 & French Guiana \\ 
\midrule
\multirow{4}{*}{\rotatebox{90}{\scriptsize Additional}} 
& Wytham woods~\cite{calders2022laser} &
  Deciduous temperate forest (\textit{Fraxinus excelsior}, \textit{Acer pseudoplatanus}, \textit{Corylus avellana}) &
  TLS (RIEGL VZ-400) &
  \phantom{0}0 & \phantom{0}0 & \phantom{0}1 & \phantom{0}835 & United Kingdom\\
& LAUTx~\cite{tockner_andreas_2022_6560112} &
  Mixed temperate broad leaved, coniferous forest (Mixed species) &
  MLS (GeoSLAM ZEB-HORIZON) &
  \phantom{0}0 & \phantom{0}0 & \phantom{0}6 & \phantom{0}516 & Austria \\ 
\bottomrule
\end{tabular}
\end{table*}

\section{Comparison to baselines across nine regions in the FOR-instanceV2 test split} \label{sec:OtherComparisontobaselines}


We evaluate ForestFormer3D's quantitative performance across nine different forest regions in the FOR-instanceV2 test set and compare it against other methods (see \cref{tab:comparedifferentregionswithsemantic}). It is important to note that CHM-based YOLOv5~\cite{Straker2023Instance}, ForAINet~\cite{Xiang2024Automated}, and SegmentAnyTree~\cite{WIELGOSZ2024114367} are trained on the original FOR-instance dataset~\cite{Puliti2023Forinstance}, which does not include data from the NIBIO\_MLS, BlueCat, and Yuchen regions. As shown in \cref{tab:comparedifferentregionswithsemantic}, ForestFormer3D consistently achieves the best results for individual tree segmentation across all nine regions except for SCION. In simpler forest areas, such as CULS, NIBIO, SCION, and NIBIO\_MLS, ForestFormer3D maintains high F1 and Cov scores. In more challenging environments, such as TUWIEN (where trees are closely intertwined), RMIT and Yuchen (with sparse point densities), and NIBIO2 and BlueCat (characterized by both large canopy trees and closely packed understory trees), ForestFormer3D surpasses all prior baselines. It demonstrates substantial improvements in regions that were previously difficult for existing methods. For example, in the RMIT region, ForestFormer3D improves the F1 by 6.8\,pp over the second best one. In the SCION region, the Cov metric shows an increase of 5.1\,pp. The NIBIO2 region is particularly challenging due to the presence of both understory and canopy trees, as noted in prior work~\cite{Xiang2024Automated}. Despite this, ForestFormer3D outperforms the OneFormer3D baseline by 8.2\,pp in F1 and achieves a 17.6\,pp improvement over the reported F1 of 72.8\% from ForAINet~\cite{Xiang2024Automated}. 

For semantic segmentation, ForestFormer3D demonstrates competitive or superior performance in most regions, with slight decreases in mIoU observed only in the CULS and NIBIO\_MLS regions compared to the best-performing method.

\begin{table*}[htbp]
\centering
\caption{Comparison with other baselines on the individual tree segmentation and semantic segmentation for different forest regions in FOR-instanceV2 test split. The best results are in bold, and the second best ones are \underline{underlined}.}
\label{tab:comparedifferentregionswithsemantic}
\scriptsize
\begin{tabular}{@{}p{1cm} p{2.8cm} p{0.6cm} p{0.6cm} p{0.6cm} p{0.6cm} p{0.6cm} p{0.6cm} p{0.6cm} p{0.6cm}@{}}
\toprule
\multirow{2}{*}{Region} & \multirow{2}{*}{Method} & \multicolumn{4}{c}{Individual Tree Seg. (\%)} & \multicolumn{4}{c}{Semantic Seg. (\%)} \\
\cmidrule(l){3-6} \cmidrule(l){7-10} 
                       &                     & \phantom{0}Prec & \phantom{0}Rec & \phantom{0}F1 & Cov & Ground & Wood & Leaf & mIoU \\
\midrule
\multirow{6}{*}{CULS}  & CHM-based YOLOv5~\cite{Straker2023Instance}       & \textbf{100.0} & \textbf{100.0} & \textbf{100.0} & -- & --  & -- & -- & -- \\
& ForAINet~\cite{Xiang2024Automated}       &   \phantom{0}87.0 & \textbf{100.0} & \phantom{0}93.0 & \underline{98.2} & -- & -- & -- & --  \\
& SegmentAnyTree~\cite{WIELGOSZ2024114367}       & \textbf{100.0}  & \textbf{100.0} & \textbf{100.0} & -- & -- & -- & -- & -- \\
& ForAINetV2\_R16       &  \textbf{100.0}   &  \textbf{100.0}   &  \textbf{100.0}   &   96.6   &   \textbf{99.8}   &   \underline{67.2}   & \underline{96.4} & \underline{87.8}  \\
                       & OneFormer3D~\cite{Kolodiazhnyi2024OneFormer3D} &   \phantom{0}\underline{95.0}   &  \phantom{0}\underline{95.0}   &  \phantom{0}\underline{95.0}   &   94.6   &   \textbf{99.8}   &   \textbf{69.0}   & \textbf{96.5} & \textbf{88.5} \\
                       & ForestFormer3D (ours) &   \textbf{100.0}   &  \textbf{100.0}   &  \textbf{100.0}   &   \textbf{99.4}   &   \textbf{99.8}   &   66.1   & 96.2 & 87.4 \\
\midrule
\multirow{6}{*}{NIBIO} & CHM-based YOLOv5~\cite{Straker2023Instance}       &  \phantom{0}87.0 & \phantom{0}72.0 & \phantom{0}79.0 & -- & --  & -- & -- & -- \\
& ForAINet~\cite{Xiang2024Automated}       &   \phantom{0}96.4 & \phantom{0}88.4 & \phantom{0}92.4 & 79.4 & -- & -- & -- & --  \\
& SegmentAnyTree~\cite{WIELGOSZ2024114367}       & \phantom{0}91.0  & \phantom{0}88.0 & \phantom{0}89.5 & -- & -- & -- & -- & -- \\
& ForAINetV2\_R16       & \phantom{0}\textbf{98.1}   &  \phantom{0}89.4   &  \phantom{0}\underline{93.4}   &   82.4   &   \underline{97.9}   &   \underline{62.1}   & \underline{93.3} & \underline{84.4}    \\
                       & OneFormer3D~\cite{Kolodiazhnyi2024OneFormer3D} &   \phantom{0}79.2   &  \phantom{0}\textbf{96.2}   &  \phantom{0}86.6   &   \textbf{88.9}   &   97.8   &   60.7   & 93.0 & 83.9 \\
                       & ForestFormer3D (ours) &   \phantom{0}\underline{97.7}   &  \phantom{0}\underline{95.8}   &  \phantom{0}\textbf{96.7}   &   \underline{88.8}   &   \textbf{98.1}   &   \textbf{64.2}   & \textbf{93.5} & \textbf{85.2} \\
\midrule
\multirow{6}{*}{RMIT}  & CHM-based YOLOv5~\cite{Straker2023Instance}       &  \phantom{0}70.0 & \phantom{0}62.0 & \phantom{0}65.0 & -- & --  & -- & -- & -- \\
& ForAINet~\cite{Xiang2024Automated}       &   \phantom{0}75.9 & \phantom{0}64.1 & \phantom{0}69.5 & 60.6 & -- & -- & -- & -- \\
& SegmentAnyTree~\cite{WIELGOSZ2024114367}       & \phantom{0}\textbf{83.0}  & \phantom{0}69.0 & \phantom{0}\underline{75.4} & -- & -- & -- & -- & -- \\
& ForAINetV2\_R16       &    \phantom{0}74.5   &  \phantom{0}59.4   &  \phantom{0}66.1   &   60.3   &   \textbf{98.3}   &   \underline{55.8}   & \underline{92.7} & \underline{82.3}     \\
                       & OneFormer3D~\cite{Kolodiazhnyi2024OneFormer3D} &   \phantom{0}70.3   &  \phantom{0}\underline{81.2}   &  \phantom{0}\underline{75.4}   &   \textbf{73.7}   &   98.1   &   52.7   & \textbf{92.8} & 81.2 \\
                       & ForestFormer3D (ours) &   \phantom{0}\underline{81.5}   &  \phantom{0}\textbf{82.8}   &  \phantom{0}\textbf{82.2}   &   \underline{73.5}   &   \underline{98.2}   &   \textbf{58.1}   & \underline{92.7} & \textbf{83.0} \\
\midrule
\multirow{6}{*}{SCION} & CHM-based YOLOv5~\cite{Straker2023Instance}       &  \phantom{0}91.0 & \phantom{0}91.0 & \phantom{0}91.0 & -- & --  & -- & -- & -- \\
& ForAINet~\cite{Xiang2024Automated}       &   \phantom{0}96.0 & \phantom{0}87.7 & \phantom{0}91.5 & 83.1 & -- & -- & -- & -- \\
& SegmentAnyTree~\cite{WIELGOSZ2024114367}       & \phantom{0}93.0  & \phantom{0}\underline{92.0} & \phantom{0}92.5 & -- & -- & -- & -- & -- \\
& ForAINetV2\_R16       &  \textbf{100.0}   &  \phantom{0}90.4   &  \phantom{0}\textbf{95.0}   &   83.2   &   \textbf{99.7}   &   \underline{61.9}   & \underline{95.1} & \underline{85.6}     \\
                       & OneFormer3D~\cite{Kolodiazhnyi2024OneFormer3D} &   \phantom{0}58.7   &  \phantom{0}\textbf{92.4}   &  \phantom{0}71.7   &   \underline{83.6}   &   \textbf{99.7}   &   61.7   & 95.0 & 85.5 \\
                       & ForestFormer3D (ours) &   \phantom{0}\underline{97.1}   &  \phantom{0}\textbf{92.4}   &  \phantom{0}\underline{94.7}   &   \textbf{88.7}   &   \textbf{99.7}   &   \textbf{64.4}   & \textbf{95.5} & \textbf{86.5} \\
\midrule
\multirow{6}{*}{TUWIEN} & CHM-based YOLOv5~\cite{Straker2023Instance} & \phantom{0}41.0 & \phantom{0}23.0 & \phantom{0}30.0 & -- & -- & -- & -- & -- \\
& ForAINet~\cite{Xiang2024Automated} & \phantom{0}66.6 & \phantom{0}\textbf{71.4} & \phantom{0}\underline{69.4} & \textbf{58.3} & -- & -- & -- & -- \\
& SegmentAnyTree~\cite{WIELGOSZ2024114367} & \phantom{0}55.0 & \phantom{0}46.0 & \phantom{0}50.1 & -- & -- & -- & -- & -- \\
& ForAINetV2\_R16 & \phantom{0}\underline{68.2} & \phantom{0}42.9 & \phantom{0}52.6 & 47.9 & 98.2 & \textbf{53.3} & \textbf{94.5} & \textbf{82.0} \\
& OneFormer3D~\cite{Kolodiazhnyi2024OneFormer3D} & \phantom{0}41.5 & \phantom{0}62.9 & \phantom{0}50.0 & \underline{54.8} & \underline{98.4} & 48.4 & \underline{94.2} & 80.3 \\
& ForestFormer3D (ours) & \phantom{0}\textbf{92.0} & \phantom{0}\underline{65.7} & \phantom{0}\textbf{76.7} & \underline{54.8} & \textbf{98.5} & \underline{52.8} & \underline{94.2} & \underline{81.8} \\
\midrule
\multirow{4}{*}{NIBIO2} & ForAINet~\cite{Xiang2024Automated}       &  \phantom{0}83.5  & \phantom{0}64.5& \phantom{0}72.8 & --  & -- & -- & -- \\
& ForAINetV2\_R16       & \phantom{0}\underline{91.8}   &  \phantom{0}72.9   &  \phantom{0}80.6   &   70.0   &   \underline{96.5}   &   52.9   & 95.7 & 81.7       \\
                       & OneFormer3D~\cite{Kolodiazhnyi2024OneFormer3D} &   \phantom{0}79.2   &  \phantom{0}\underline{85.9}   &  \phantom{0}\underline{82.2}   &   \underline{79.1}   &   \textbf{96.9}   &   \underline{54.4}   & \underline{95.8} & \underline{82.3} \\
                       & ForestFormer3D (ours) &   \phantom{0}\textbf{94.6}   &  \phantom{0}\textbf{86.9}   &  \phantom{0}\textbf{90.4}   &   \textbf{80.2}   &   \textbf{96.9}   &   \textbf{56.6}   & \textbf{95.9} & \textbf{83.1} \\
\midrule
\multirow{3}{*}{NIBIO\_MLS} & ForAINetV2\_R16   & \phantom{0}\underline{95.5}   &  \phantom{0}\underline{91.3}   &  \phantom{0}\underline{93.3}   &   84.4   &   \underline{98.8}   &   74.9   & 86.4 & \underline{86.7}     \\
                       & OneFormer3D~\cite{Kolodiazhnyi2024OneFormer3D} &   \phantom{0}79.3   &  \textbf{100.0}   &  \phantom{0}88.5   &   \textbf{89.8}   &   \textbf{98.9}   &   \underline{75.3}   & \textbf{87.2} & \textbf{87.1} \\
                       & ForestFormer3D (ours) &   \textbf{100.0}   &  \phantom{0}\underline{91.3}   &  \phantom{0}\textbf{95.5}   &   \underline{85.4}   &   94.5   &   \textbf{77.6}   & \underline{87.1} & 86.4 \\
\midrule
\multirow{3}{*}{BlueCat} & ForAINetV2\_R16     &      \phantom{0}\underline{71.8}   &  \phantom{0}27.9   &  \phantom{0}40.2   &   32.8   &   --   &   \textbf{73.3}   & \textbf{94.7} & \textbf{84.0}      \\
                       & OneFormer3D~\cite{Kolodiazhnyi2024OneFormer3D} &   \phantom{0}59.7   &  \phantom{0}\underline{47.1}   &  \phantom{0}\underline{52.6}   &   \underline{48.3}   &   --   &   64.4   & 93.0 & 78.7 \\
                       & ForestFormer3D (ours) &   \phantom{0}\textbf{84.5}   &  \phantom{0}\textbf{48.6}   &  \phantom{0}\textbf{61.7}   &   \textbf{48.8}   &   --   &   \underline{69.9}   & \underline{93.9} & \underline{81.9} \\
\midrule
\multirow{3}{*}{Yuchen} & ForAINetV2\_R16      &     \phantom{0}\underline{78.3}   &  \phantom{0}\underline{75.0}   &  \phantom{0}\underline{76.6}   &   \underline{74.9}   &   99.4   &   \textbf{51.7}   & \textbf{98.1} & \textbf{83.0}     \\
                       & OneFormer3D~\cite{Kolodiazhnyi2024OneFormer3D} &   \phantom{0}52.9   &  \phantom{0}\underline{75.0}   &  \phantom{0}62.1   &   71.6   &   \underline{99.6}   &   \underline{50.0}   & \underline{97.9} & \underline{82.5} \\
                       & ForestFormer3D (ours) &   \phantom{0}\textbf{90.5}   &  \phantom{0}\textbf{79.2}   &  \phantom{0}\textbf{84.4}   &   \textbf{79.2}   &   \textbf{99.7}   &   49.7   & \textbf{98.1} & \underline{82.5} \\
\bottomrule
\end{tabular}
\end{table*}

\section{Additional ablation studies}
\label{sec:Otherablationstudies}

\begin{figure}[!htbp]
  \captionsetup{skip=0pt} 
  \centering
  \includegraphics[width=\columnwidth]{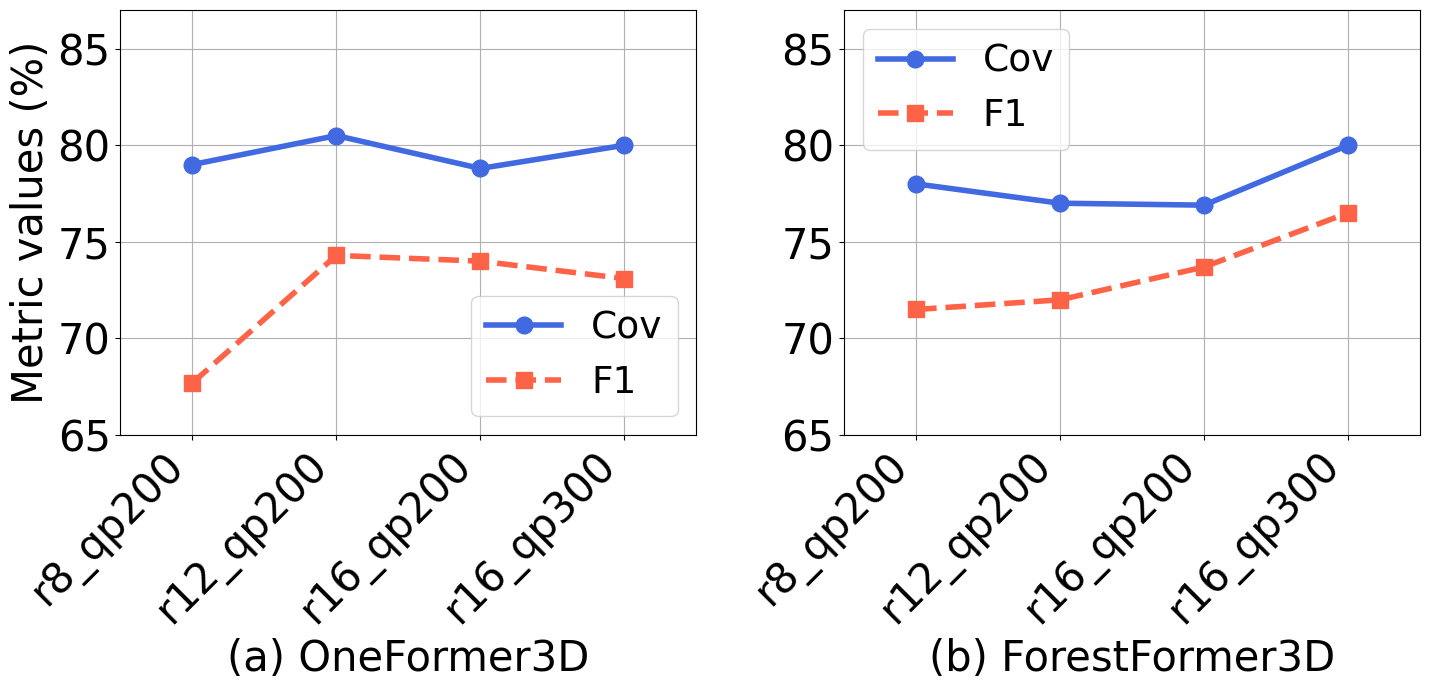}
  \caption{Impact of cylinder input radius and number of query points on individual tree segmentation metrics for FOR-instanceV2 test split.}
  \label{fig:ablationradiusandqp}
  \vspace*{0pt} 
\end{figure}

\noindent\textbf{Input radius size and number of query points.} \cref{fig:ablationradiusandqp} compares the effect of different cylinder input radii and the number of query points on individual tree segmentation metrics (F1 and Cov) within the FOR-instanceV2 test split. \cref{fig:ablationradiusandqp}(a) shows the performance of OneFormer3D, while \cref{fig:ablationradiusandqp}(b) shows the performance of ForestFormer3D. \texttt{r8\_qp200} represents a cylinder radius of 8\,\text{m} with 200 query points, with similar interpretations for other settings. For ForestFormer3D, increasing the input radius and the number of query points generally results in higher F1 scores. However, due to GPU memory limitations, further increases in these parameters were not explored. For OneFormer3D, achieving an optimal balance between the input radius and the number of query points is crucial to better performance.

\begin{figure}[!htbp]
  \captionsetup{skip=0pt} 
  \centering
  \includegraphics[width=\columnwidth]{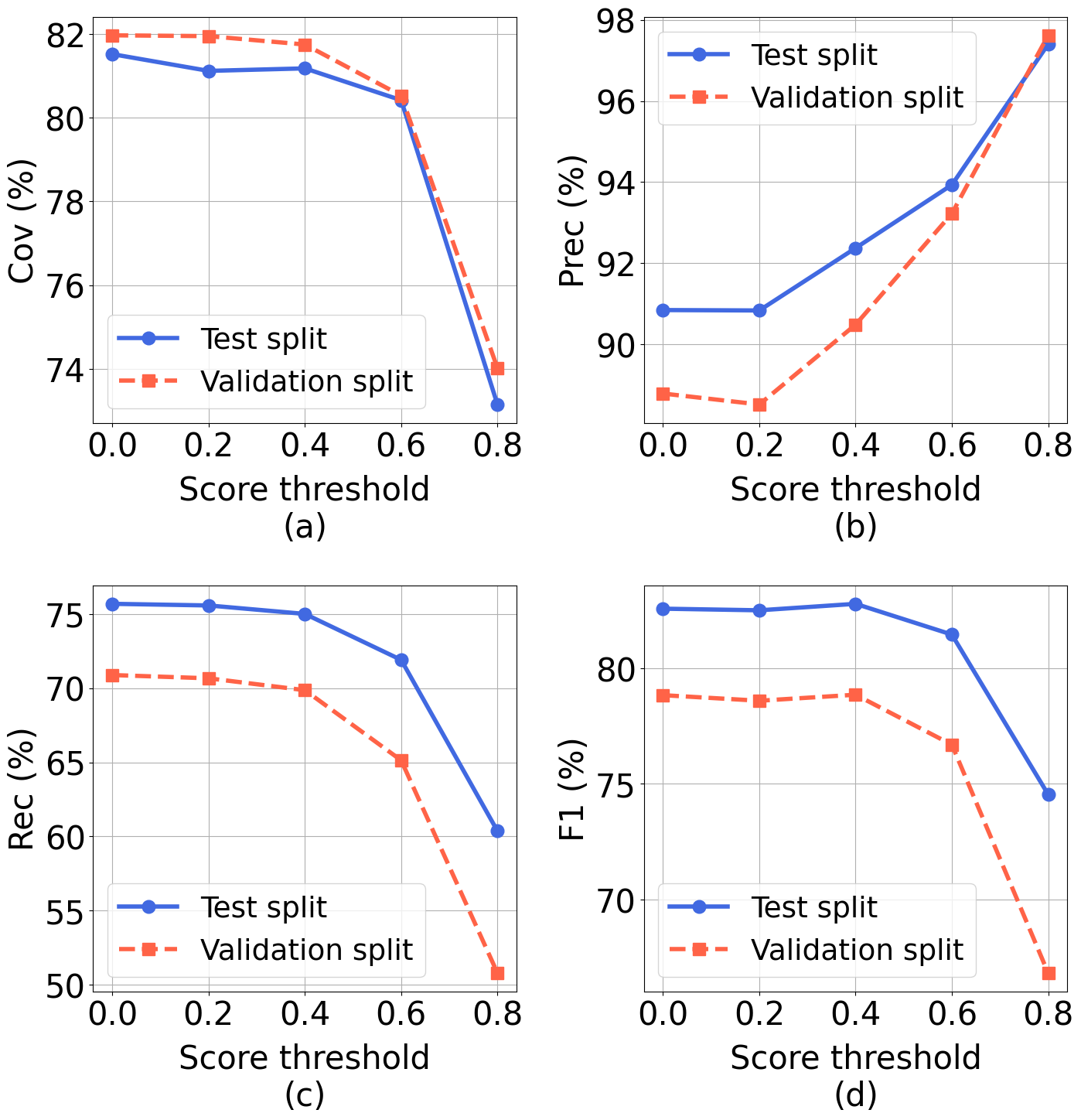}
  \caption{Effect of score threshold on key metrics of individual tree segmentation (Prec, Rec, F1 and Cov) for FOR-instanceV2 dataset.}
  \label{fig:ablationscorethreshold}
  \vspace*{0pt} 
\end{figure}

\noindent\textbf{Score threshold on mask removal.} We evaluate the effect of the score threshold on Prec, Rec, F1, and Cov using the FOR-instanceV2 dataset (see \cref{fig:ablationscorethreshold}). During inference, masks with scores below the threshold are removed as unreliable, affecting the balance between retaining true positives and removing false positives. As shown in \cref{fig:ablationscorethreshold}, higher thresholds improve Prec (\cref{fig:ablationscorethreshold}(b)) by reducing false positives but decrease Cov (\cref{fig:ablationscorethreshold}(a)) and Rec (\cref{fig:ablationscorethreshold}(c)) by discarding true positives. \cref{fig:ablationscorethreshold}(d) highlights the trade-off between Prec and Rec. A threshold of 0.4 achieves the best F1 score in both validation and test sets, making it the optimal choice for our framework.

\begin{figure}[!htbp]
  \captionsetup{skip=0pt} 
  \centering
  \includegraphics[width=\columnwidth]{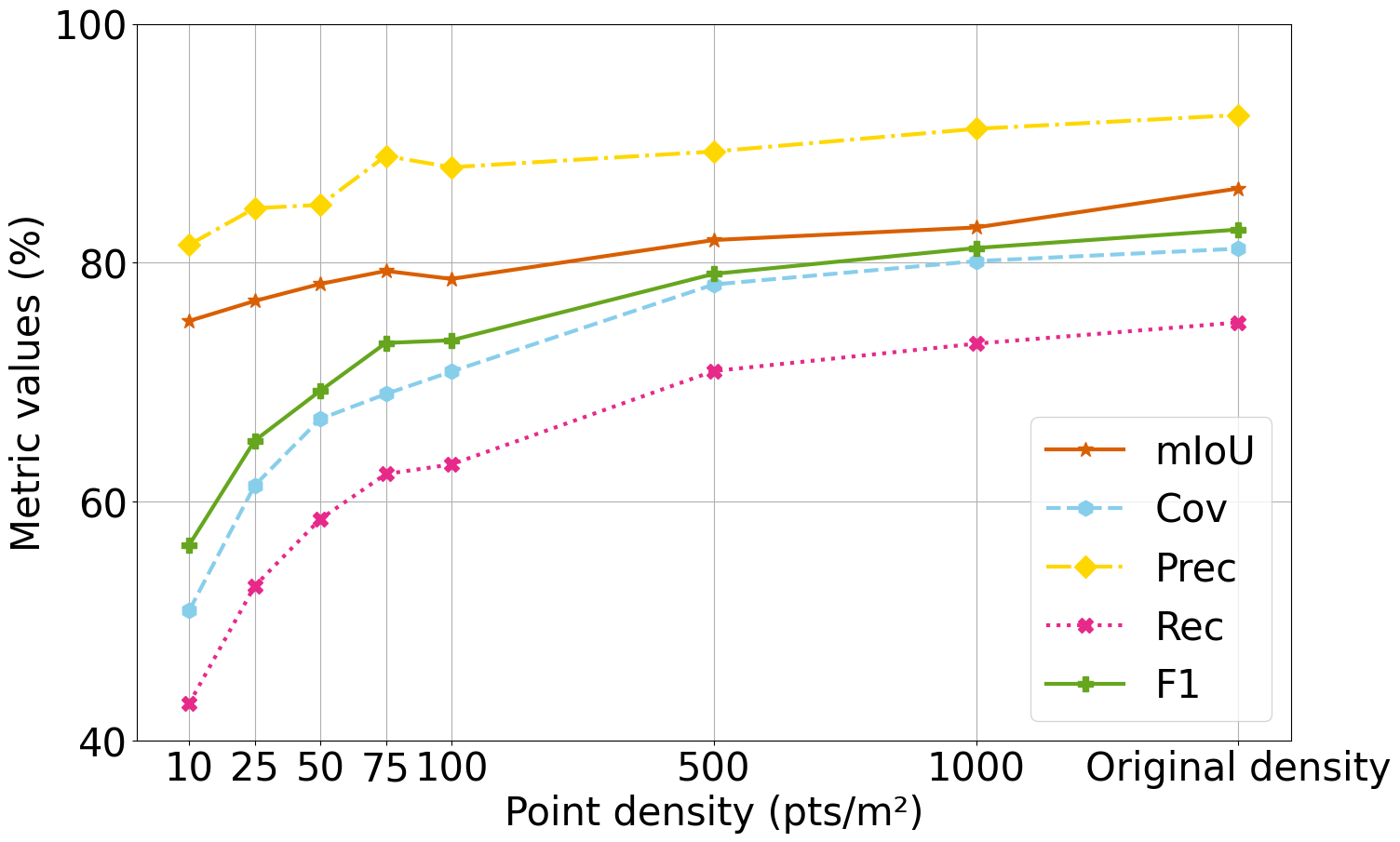}
  \caption{Effect of point density on metrics of individual tree segmentation (Prec, Rec, F1 and Cov) and semantic segmentation (mIoU) for FOR-instanceV2 test split.}
  \label{fig:ablationpointdensity}
  \vspace*{-10pt} 
\end{figure}

\noindent\textbf{Point cloud density.} We evaluated the impact of different point cloud densities on segmentation performance. The original FOR-instanceV2 dataset was downsampled to seven densities: 10, 25, 50, 75, 100, 500, and 1000 points per square meter (pts/m²). The results are shown in \cref{fig:ablationpointdensity}. As the point density decreases, all metrics experience varying degrees of degradation. In particular, mIoU and Prec remain relatively stable in different densities, while Rec, Cov, and F1 show more noticeable drops when the density falls below 500\,pts/m². This suggests that the method may face certain limitations in extremely sparse point cloud scenarios.

\begin{figure}[!htbp]
  \captionsetup{skip=0pt} 
  \centering
  \includegraphics[width=\columnwidth]{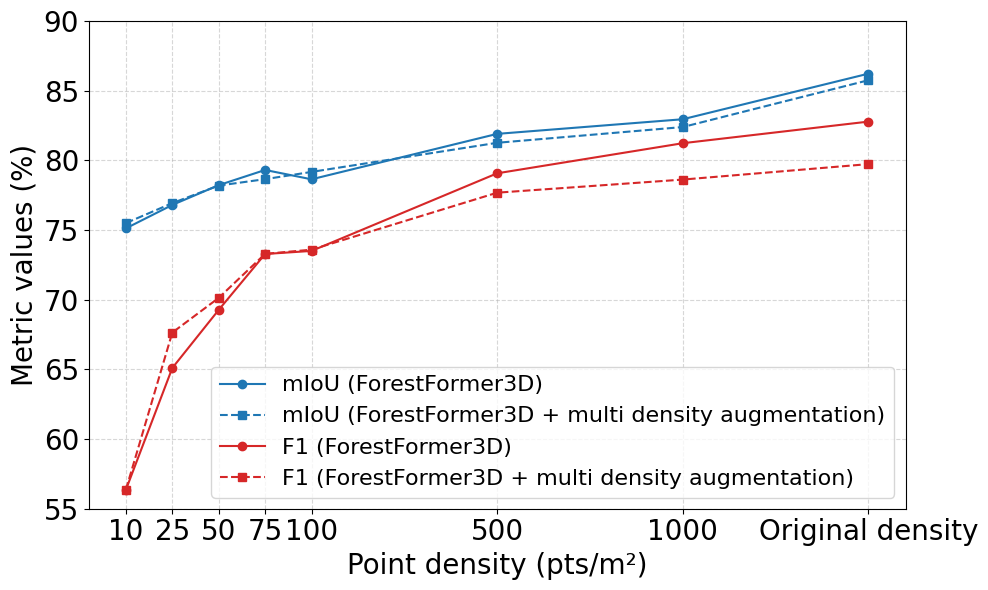}
  \caption{Effect of multi density augmentation on individual tree segmentation (F1) and semantic segmentation (mIoU) on the FOR-instanceV2 test split.}
  \label{fig:ablationpointdensity2}
  \vspace*{-10pt} 
\end{figure}

To improve robustness under varying point cloud densities, we retrained ForestFormer3D using the multi density augmentation strategy from SegmentAnyTree~\cite{WIELGOSZ2024114367}. The training set combined point clouds from eight densities (i.e., the original resolution and seven downsampled versions). As shown in~\cref{fig:ablationpointdensity2}, this strategy improves individual tree segmentation under sparse conditions (e.g.,~$<\!100$\,pts/m²), yielding higher F1 scores. However, it brings no benefit for dense inputs and can even degrade performance. These results suggest that simple multi-density training is insufficient to address the challenges posed by point density variation, which remains an open problem.

\section{Qualitative results}
\label{sec:Qualitativeresults}
To provide an intuitive understanding of how segmentation metrics relate to actual segmentation quality, we present visualizations of ForestFormer3D's results in individual tree segmentation and semantic segmentation tasks.

\begin{figure*}[]
  \captionsetup{skip=0pt} 
  \centering
  \includegraphics[width=0.8\textwidth]{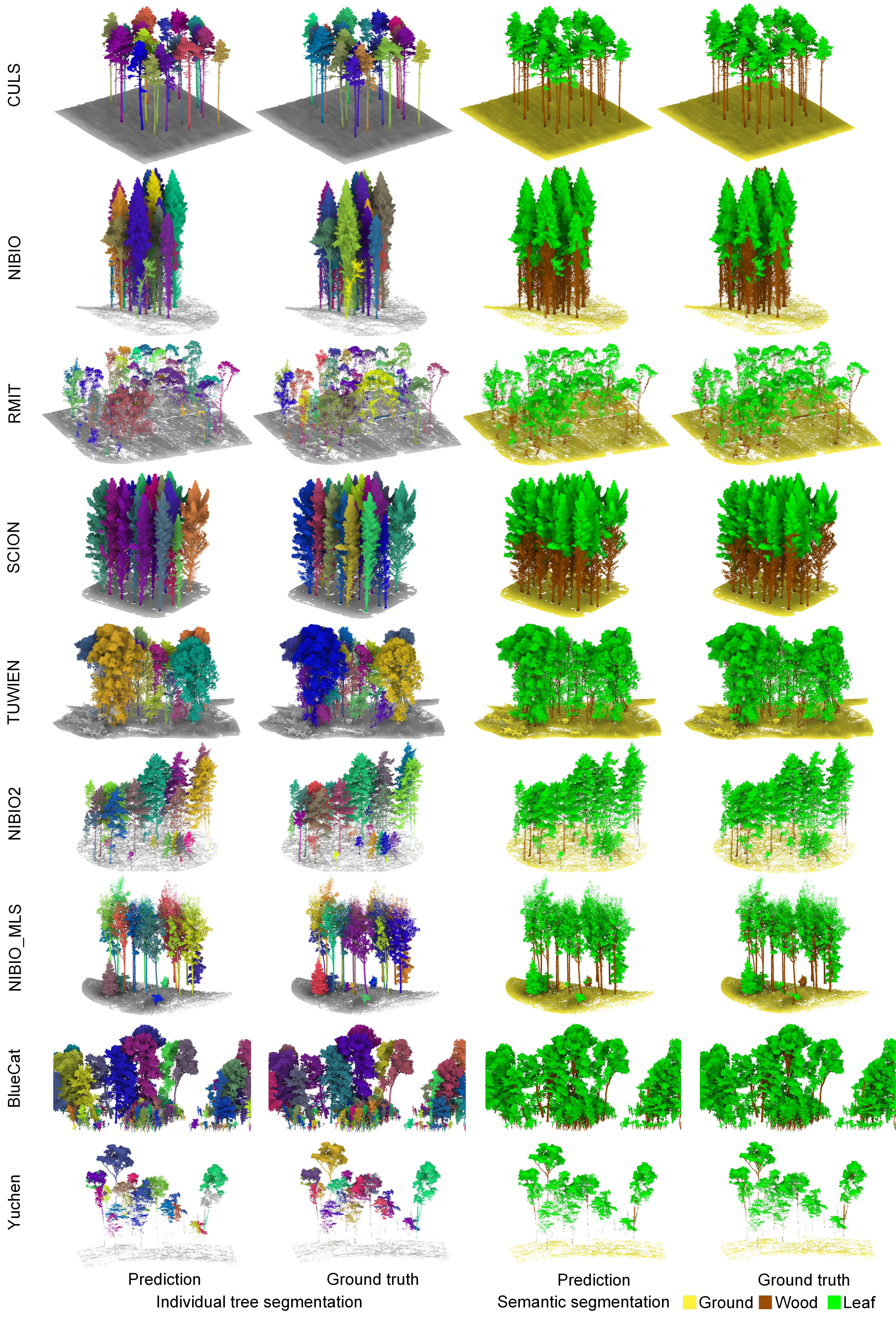}
  \caption{Visualization of ForestFormer3D segmentation results for nine different regions in FOR-instanceV2 test split. The colors for the individual trees are assigned randomly.}
  \label{fig:visulizationfor9regions}
  \vspace*{-10pt} 
\end{figure*}

\cref{fig:visulizationfor9regions} shows both individual tree segmentation and semantic segmentation predictions, along with corresponding ground truth segmentations, across nine different forest regions in the FOR-instanceV2 test split. The results show that our ForestFormer3D performs well under various conditions, including data collected from different sensors, different point densities, varying forest types and geographical locations (\ie, boreal, temperate and tropical forests), and complex environments where both large and small trees coexist. These visual results demonstrate the robustness of ForestFormer3D in handling diverse and challenging forest scenarios.

\begin{figure*}[]
  \captionsetup{skip=0pt} 
  \centering
  \includegraphics[width=\textwidth]{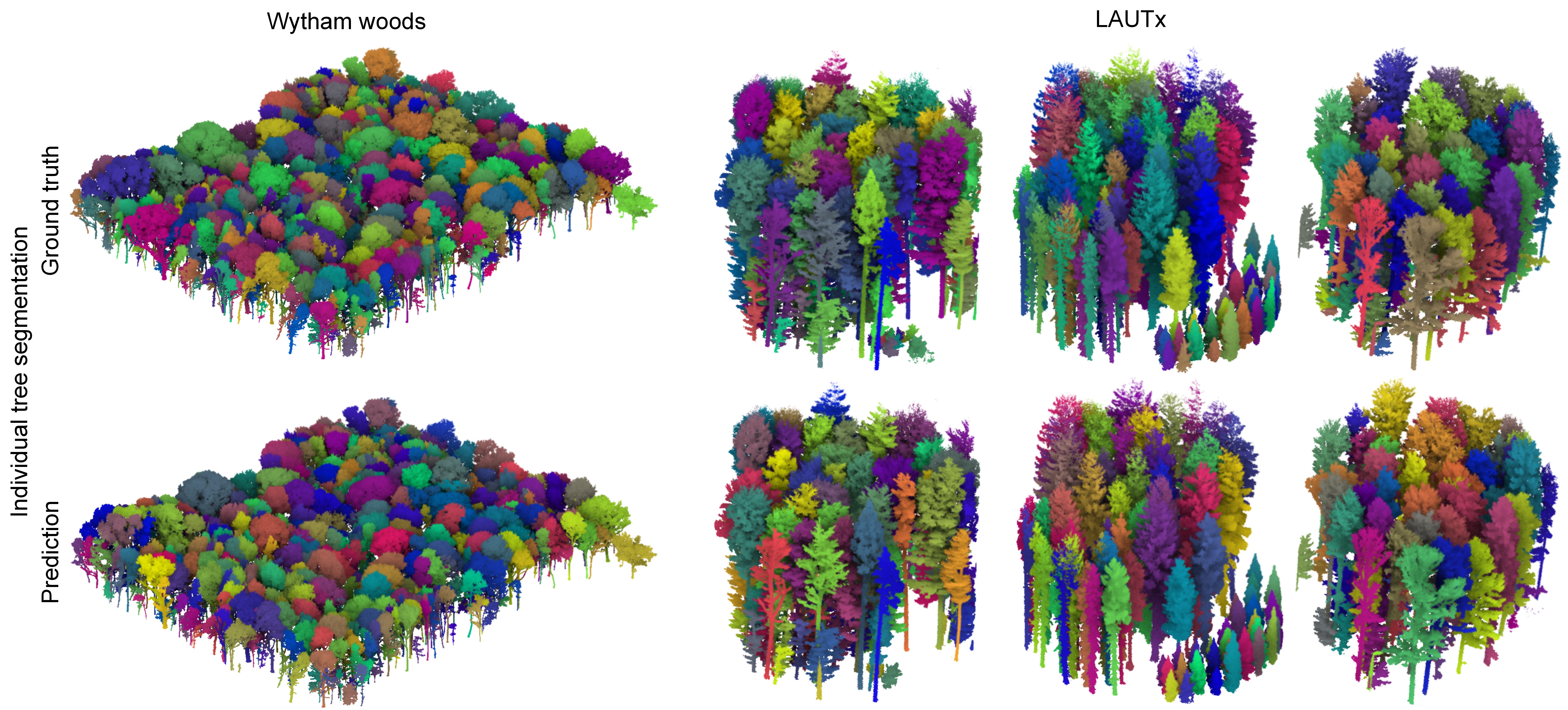}
  \caption{Visualization of ForestFormer3D individual tree segmentation results for Wytham woods and LAUTx. The colors for the individual trees are assigned randomly.}
  \label{fig:visulizationforotherregions}
\end{figure*}

\begin{figure*}[]
  \captionsetup{skip=0pt} 
  \centering
  \includegraphics[width=0.8\textwidth]{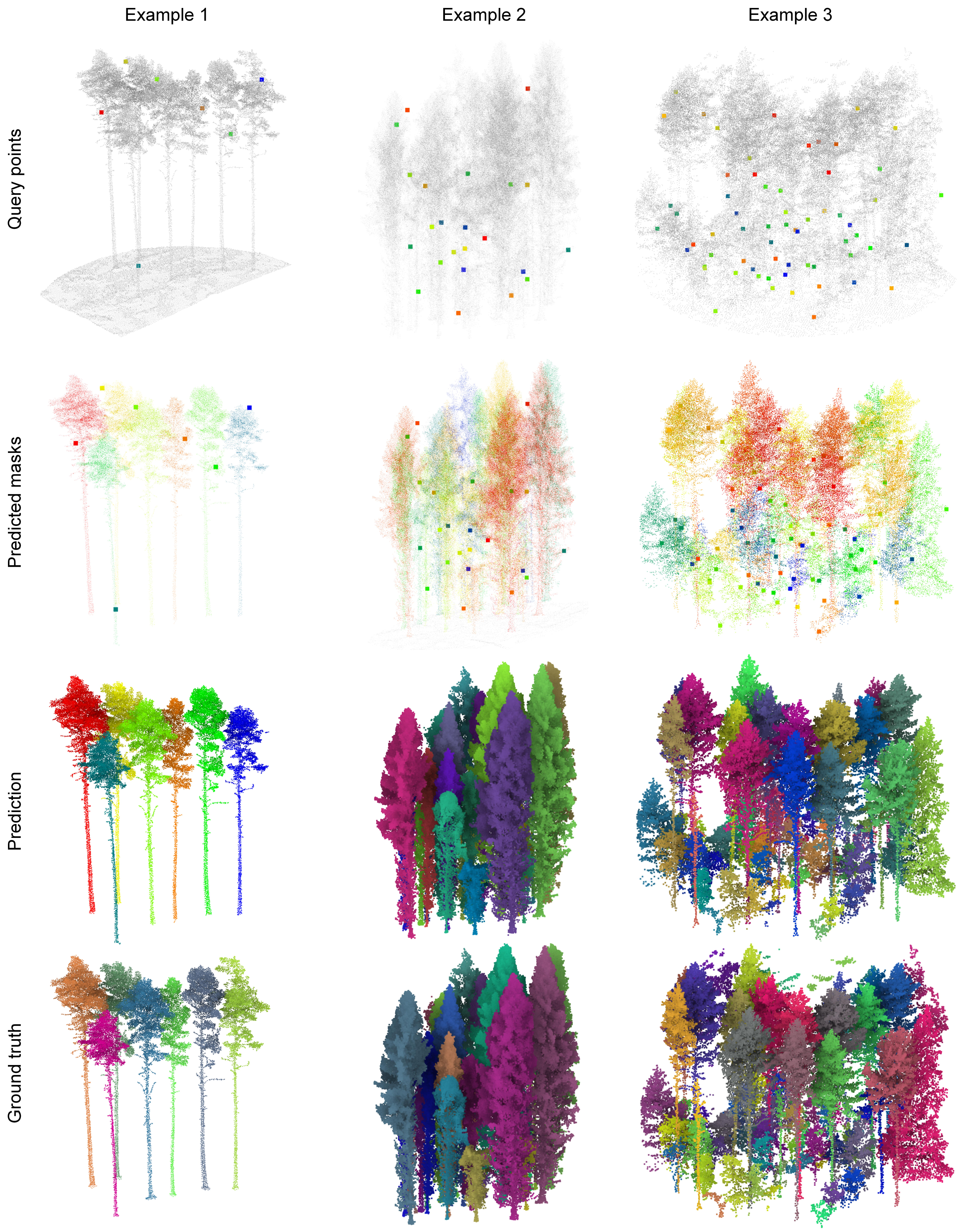}
  \caption{Visual comparison of individual tree segmentation for three representative forest scenes, with each row illustrating a different stage in our segmentation framework. In the first row, the selected ISA-guided query points are highlighted as enlarged, randomly colored points, effectively covering nearly every tree in the scene. The second row shows the predicted masks generated based on the query points from the first row, with each mask color corresponding to its associated query point color. The third row increases color contrast to distinguish each predicted individual tree, with colors again randomly assigned. The fourth row presents the ground truth segmentation with randomly assigned colors to aid comparison.} \label{fig:visulizationforqueryselection}
\end{figure*}

\cref{fig:visulizationforotherregions} displays individual tree segmentation results with predicted and ground truth segmentations for ForestFormer3D on the Wytham woods and LAUTx datasets. \cref{fig:visulizationforqueryselection} presents three representative examples, progressing from simple forest scenes to densely packed environments with both large canopy trees and understory ones. For each example, we include visualizations of the ISA-guided query points, the predicted masks generated from these points, and the final individual tree segmentation predictions, along with the ground truth for comparison.

\begin{figure*}[]
  \captionsetup{skip=0pt} 
  \centering
  \includegraphics[width=0.9\textwidth]{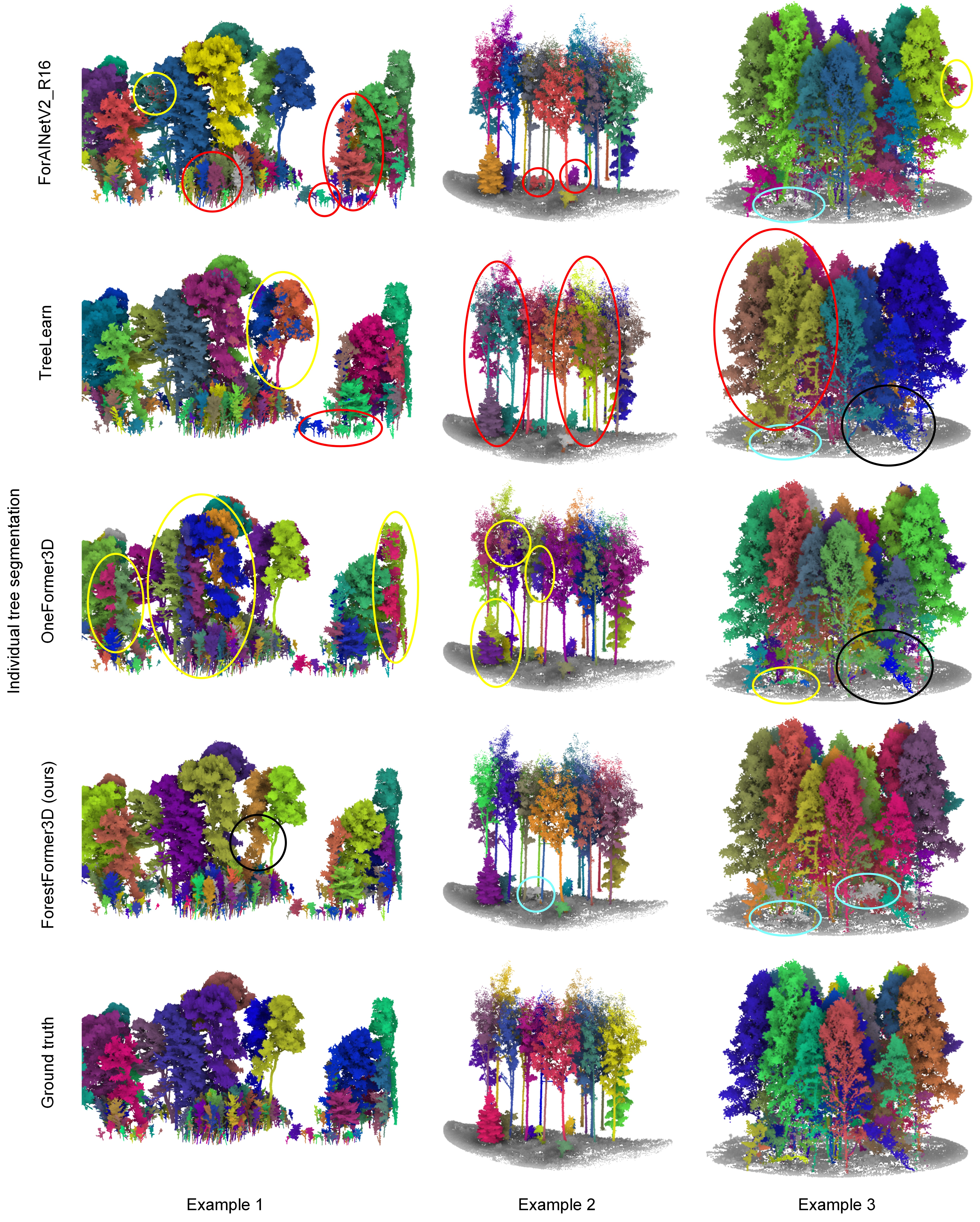}
  \caption{Visual comparison of individual tree segmentation results among ForestFormer3D and baseline methods, with randomly assigned colors representing different tree instances. Common segmentation errors are highlighted using four distinct markers: red circles for under-segmentation, yellow circles for over-segmentation, bright blue circles for undetected trees, and black circles for other errors, such as points from one tree being assigned to a neighboring tree.} \label{fig:comparewithotherbaselines}
\end{figure*}

\cref{fig:comparewithotherbaselines} presents a visual comparison of ForestFormer3D and baseline methods. In Example 1 and Example 2, ForestFormer3D demonstrates its ability to effectively mitigate over-segmentation and under-segmentation compared to other methods. However, in Example 2, small trees are occasionally missed, indicating a potential limitation in detecting such cases. In Example 3, heavily inclined small trees near the ground pose significant challenges for all methods, leading to either missed detections, over-segmentation, or incorrect assignment to neighboring larger trees. Addressing these complex scenarios remains an open problem for future research.

\section{Relation to forest domain specific metrics} \label{sec:MetricRelation}

While domain-specific metrics have long been used in forestry research, we adopt Prec, Rec, and F1 to align with standard evaluation practices in computer vision. This allows for direct comparability across domains while maintaining consistency with traditional forestry metrics. Specifically, completeness corresponds to Rec, omission error to \( 1 - \text{Rec} \), commission error to \( 1 - \text{Prec} \), and F-score remains equivalent to F1. These relationships have been previously established in forestry research~\cite{Xiang2024Automated}, ensuring that our reported results remain interpretable for both forestry and computer vision communities.

It is worth noting that SegmentAnyTree~\cite{WIELGOSZ2024114367} reports locally computed F1 scores, whereas we adopt global metrics consistent with ForAINet~\cite{Xiang2024Automated}. This difference in evaluation metrics explains the discrepancy in reported F1 scores across methods.

{
    \small
    \bibliographystyle{ieeenat_fullname}
    \bibliography{main}
}

\end{document}